\documentclass{article} 
\usepackage{times}
\PassOptionsToPackage{numbers, compress}{natbib}

\usepackage[preprint]{neurips_2021}

\usepackage{amsmath,amsfonts,bm}

\def\eqref#1{equation~\ref{#1}}

\def\1{\bm{1}}

\DeclareMathAlphabet{\mathsfit}{\encodingdefault}{\sfdefault}{m}{sl}
\SetMathAlphabet{\mathsfit}{bold}{\encodingdefault}{\sfdefault}{bx}{n}

\usepackage{hyperref}
\hypersetup{colorlinks,linkcolor={blue},citecolor={blue},urlcolor={blue}}  

\usepackage{url}
\usepackage{graphics}
\usepackage{epsfig}
\usepackage{booktabs} 
\usepackage{amsmath}
\usepackage{subfigure}
\usepackage{multirow}
\usepackage{threeparttable}
\usepackage{caption}
\usepackage{enumitem}
\usepackage{hyperref}
\usepackage{xspace}
\usepackage{soul}
\usepackage{xcolor}

\usepackage{arydshln}

\title{\ours: A Federated Learning Benchmark System for Graph Neural Networks}

\author{Chaoyang He \& Keshav Balasubramanian \& Emir Ceyani \thanks{The first three authors contribute equally. Email: \texttt{\{chaoyang.he,keshavba,ceyani\}@usc.edu}.} \\
Viterbi School of Engineering\\
University of Southern California\\
\And
Carl Yang \& Han Xie\\
Department of Computer Science\\
Emory University
\And
Lichao Sun \& Lifang He\\
Department of Computer Science and Engineering\\
Lehigh University
\And
Liangwei Yang \& Philip S. Yu\\
Department of Computer Science\\
University of Illinois at Chicago 
\And
Yu Rong \& Peilin Zhao \& Junzhou Huang \\
Machine Learning Center \\
Tencent AI Lab \\
\And
Murali Annavaram \& Salman Avestimehr \\
Viterbi School of Engineering\\
University of Southern California\\

}

\newcommand{\ours}{\texttt{FedGraphNN}\xspace}

\newlist{myitemize}{itemize}{1}
\setlist[myitemize,1]{label=\textbullet,leftmargin=0.2in}

\newlist{myenum}{enumerate}{1}
\setlist[myenum,1]{label=\textbullet,leftmargin=0.2in}

\begin{document}

\definecolor{aurometalsaurus}{rgb}{0.43, 0.5, 0.5}

\maketitle








\begin{abstract}
Graph Neural Network (GNN) research is rapidly growing thanks to the capacity of GNNs in learning distributed representations from graph-structured data. 
However, centralizing a massive amount of real-world graph data for GNN training is prohibitive due to privacy concerns, regulation restrictions, and commercial competitions. Federated learning (FL), a trending distributed learning paradigm, provides possibilities to solve this challenge while preserving data privacy. Despite recent advances in vision and language domains, there is no suitable platform for the FL of GNNs. 
To this end, we introduce \ours, an open FL benchmark system that can facilitate research on federated GNNs. \ours is built on a unified formulation of graph FL and contains a wide range of datasets from different domains, popular GNN models, and FL algorithms, with secure and efficient system support. Particularly for the datasets, we collect, preprocess, and partition 36 datasets from 7 domains, including both publicly available ones and specifically obtained ones such as \texttt{hERG} and \texttt{Tencent}. Our empirical analysis showcases the utility of our benchmark system, while exposing significant challenges in graph FL: federated GNNs perform worse in most datasets with a non-IID split than centralized GNNs; the GNN model that attains the best result in the centralized setting may not maintain its advantage in the FL setting. These results imply that more research efforts are needed to unravel the mystery behind federated GNNs. Moreover, our system performance analysis demonstrates that the \texttt{FedGraphNN} system is computationally efficient and secure to large-scale graphs datasets. We maintain the source code at \url{https://github.com/FedML-AI/FedGraphNN}.
\end{abstract}

\section{Introduction}
\label{sec:intro} 

Graph Neural Networks (GNNs)  are state-of-the-art models that learn representations from complex graph-structured data in various domains such as drug discovery \citep{rong2020self,ddGnn1,yang2018node}, social network  \citep{graphsage,socialGnn1,He2019CascadeBGNNTE,yang2020relation}, recommendation systems \cite{wu2021fedgnn,liu2020basconv,ge2020fedner,yang2020multisage}, and traffic flow modeling \citep{trafficGnn1, trafficGnn2}.
However, due to privacy concerns, regulatory restrictions, and commercial competition, there are widespread real-world cases in which graph data is decentralized. For example, in the AI-based drug discovery industry, pharmaceutical research institutions would significantly benefit from other institutions' data, but neither can afford to disclose their private data due to commercial reasons.
 
\begin{figure}[ht]
\centering
\subfigure[Graph-level FL]{\label{fig:sub-first}\includegraphics[scale =0.24]{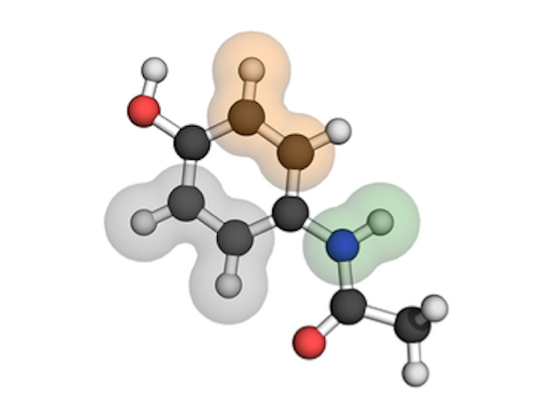}} 
\subfigure[Subgraph-level FL]{\label{fig:sub-second}\includegraphics[scale = 0.24]{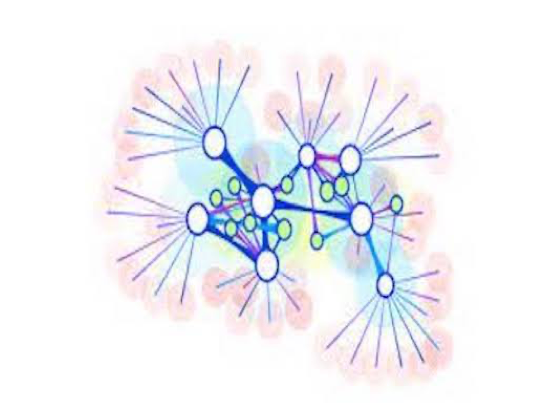}} 
\subfigure[Node-level FL]{\label{fig:sub-third}\includegraphics[scale = 0.15]{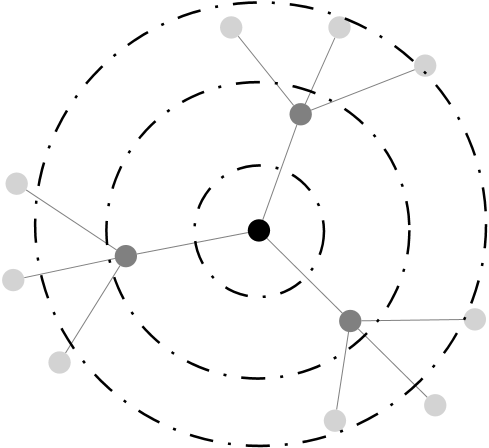}} 
\caption{Three settings of graph federated learning.} 
\label{fig:use-cases}
\end{figure}

Federated Learning (FL) is a distributed learning paradigm that addresses this data isolation problem. In FL, training is an act of collaboration between multiple clients without requiring centralized local data   \citep{mcmahan2017communication, kairouz2019advances}.
Despite its successful application in domains like computer vision \citep{he2020group,liu2020fedvision,hsu2020federated,hefedcv} and natural language processing \citep{hard2018federated, ge2020fedner,lin2021fednlp}, FL has yet to be widely adopted in the domain of machine learning on graph data. There are multiple reasons for this: 
\begin{enumerate}[leftmargin=0.5cm]
    \item There is a lack of unified formulation over the various graph FL settings and tasks in current literature, making it difficult for researchers who focus on SGD-based federated optimization algorithms to understand essential challenges in federated GNNs;
    \item Existing FL libraries, as summarized by \citep{chaoyanghe2020fedml}, do not support diverse datasets and learning tasks to benchmark different models and training algorithms. Given the complexity of graph data, the dynamics of training GNNs in an FL setting may be different from training vision or language models. A fair and easy-to-use benchmark with standardized open datasets and reference implementations is essential to the development of new graph FL models and algorithms;
    \item The simulation-oriented federated training system is inefficient and unsecure for federated GNNs research on large-scale and private graph datasets in the cross-silo settings. Disruptive research ideas may be constrained by the lack of a modularized federated training system tailored for diverse GNN models and FL algorithms. 
    
\end{enumerate}
\vspace{-0.2cm}

To address these issues, we present an open FL benchmark system for GNNs, called \ours, which contains a variety of graph datasets from different domains and eases the training and evaluation of various GNN models and FL algorithms. We first formulate graph FL to provide a unified framework for federated GNNs (Section \ref{sec:baseline}). Under this formulation, we introduce the various graph datasets with synthesized partitions according to real-world application scenarios (Section \ref{sec:datasets}). An efficient and secure FL system is designed and implemented to support popular GNN models and FL algorithms and provide low-level programmable APIs for customized research and industrial deployment (Section \ref{sec:design}). 
Extensive empirical analysis demonstrates the utility and efficiency of our system and indicates the need of further research in graph FL (Section \ref{sec:experiments}). Finally, we summarize the open challenges in graph FL based on emerging related works (Section \ref{sec:related_works}) as well as future directions based on \ours (Section \ref{sec:conclusion}).

\vspace{-5pt}
\section{Federated Graph Neural Networks (\ours)}
\label{sec:baseline}
\vspace{-5pt}

We consider a \textit{distributed graph scenario} in which a single graph is partitioned or multiple graphs are dispersed over multiple edge servers that cannot be centralized for training due to privacy or regulatory restrictions. However, collaborative training over the dispersed data can aid the formulation of more powerful and generalizable graph models. In this work, we focus on graph neural networks (GNNs) as the graph models and extend the emerging studies on federated learning (FL) over neural network models in other domains to GNNs. 

\begin{figure*}[htb!]
\centering
{\includegraphics[scale = 0.34]{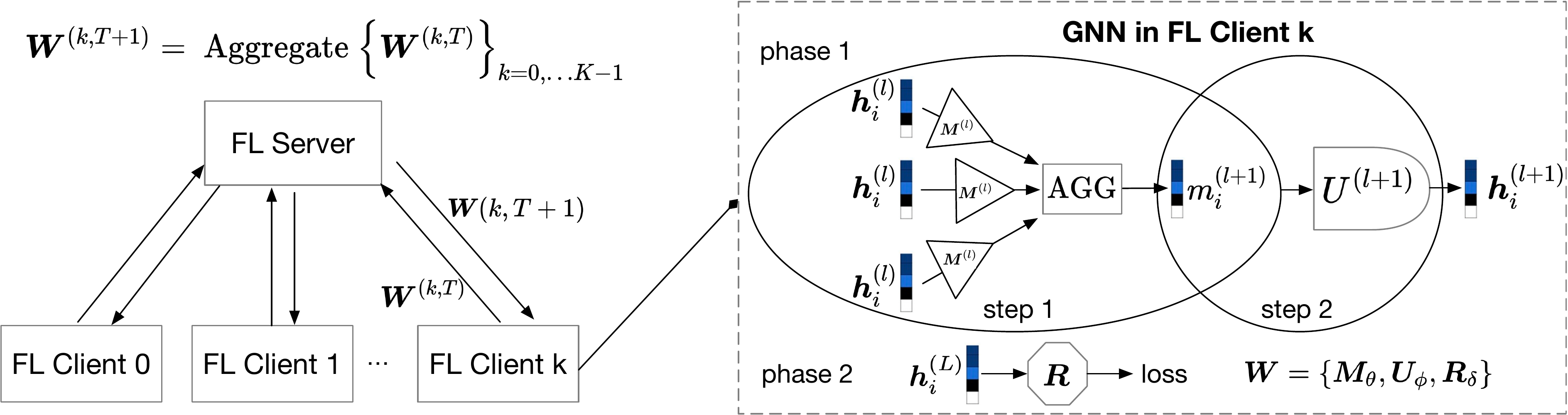}}
\caption{Formulation of \texttt{FedGraphNN} (Federated Graph Neural Network)}
\label{fig:fgnn}
\end{figure*}

In our unified framework of \ours, we assume that there are $K$ clients in the distributed graph scenario, and the $k^{th}$ client has its own dataset $\mathcal{D}^{(k)}:=(\mathcal{G}^{(k)}, \mathbf{Y}^{(k)})$, where $\mathcal{G}^{(k)}=(\mathcal{V}^{(k)}, \mathcal{E}^{(k)})$ is the graph(s) in $\mathcal{D}^{(k)}$ with vertex and edge feature sets $\boldsymbol{X}^{(k)} = \{ \boldsymbol{x}_{m}^{(k)}\}_{m \in \mathcal{V}^{(k)}}$ and $\boldsymbol{Z}^{(k)} = \{\boldsymbol{e}_{m, n}^{(k)}\}_{m, n \in \mathcal{V}^{(k)}}$, $\mathbf{Y}^{(k)}$ is the label set of $\mathcal{G}^{(k)}$. Each client owns a GNN model to learn graph representations and make predictions. Multiple clients are interested in collaborating through a server to improve their GNN models without necessarily revealing their graph datasets. 

We illustrate the formulation of \ours in Figure \ref{fig:fgnn}. 
Without loss of generality, we use a Message Passing Neural Network (MPNN) framework \citep{gilmer2017neural,10.1145/3394486.3406474}. Most of the spatial-based GNN models \citep{kipfgcn,gat,graphsage} can be unified into this framework, where the forward pass has two phases: a message-passing phase and a readout phase. 
\paragraph{GNN phase 1: Message-passing (same for all tasks).}
The message-passing phase contains two steps: (1) the model gathers and transforms the neighbors' messages, and (2) the model uses aggregated messages to update the nodes' hidden states. 
Mathematically, for client $k$ and layer indices $\ell = 0,\dots , L-1 $, an $L$-layer MPNN is formalized as follows:
\begin{gather}
\boldsymbol{m}_{i}^{(k, \ell +1)} =\operatorname{\texttt{AGG}}\left(\left\{\boldsymbol{M}_{\theta}^{(k, \ell +1)}\left(\boldsymbol{h}_{i}^{(k, \ell)}, \boldsymbol{h}_{j}^{(k, \ell)}, \boldsymbol{z}_{i, j}\right) \mid j \in \mathcal{N}_{i}\right\}\right),\nonumber\\
\boldsymbol{h}_{i}^{(k, \ell +1)} =\boldsymbol{U}_{\phi}^{(k, \ell +1)}\left(\boldsymbol{h}_{i}^{(k, \ell)}, \boldsymbol{m}_{i}^{(k, \ell +1)}\right), 
\label{eq:message}
\end{gather}
where $\boldsymbol{h}^{(k, 0)}_i = \boldsymbol{x}_{i}^{(k)}$ is the $k^{th}$ client's node features, $\ell$ is the layer index, $\texttt{AGG}$ is the aggregation function (e.g., in the GCN model, the aggregation function is a simple $\texttt{SUM}$ operation), $\mathcal{N}_{i}$ is the neighbor set of node $i$, and $\boldsymbol{M}_{\theta}^{(k, \ell+1)}\left( \cdot \right)$ is the message generation function which takes the hidden state of current node $\boldsymbol{h}_{i}$, the hidden state of the neighbor node $\boldsymbol{h}_{j}$ and the edge features $\boldsymbol{z}_{i, j}$ as inputs.  $\boldsymbol{U}_{\phi}^{(k, \ell+1)}\left( \cdot \right)$ is the state update function receiving the aggregated feature $\boldsymbol{m}_{i}^{(k, \ell+1)}$. 

\paragraph{GNN phase 2: Readout (different across tasks).}
After propagating through an $L$-layer MPNN, the readout phase computes feature vectors from the hidden states of the last MPNN layer and makes predictions for downstream tasks, that is 
\begin{align}
\hat{y}_{S}^{(k)}=\boldsymbol{R}_{\delta}\left(\left\{h_{i}^{(k, L)} \mid i \in \mathcal{V}_{S}^{(k)}\right\}\right).
\label{eq:readout}
\end{align}
Note that to handle different downstream tasks, $S$ can be a single node (node classification), a node pair (link prediction), a node set (graph classification) and so forth, and $\boldsymbol{R}_{\delta}$ can be the concatenation function or a pooling function such as $\texttt{SUM}$ plus a single- or multi-layer perceptron.

\paragraph{GNN with FL.} To formulate the FL setting, we define $\boldsymbol{W} = \left\{\boldsymbol{M}_{\theta}, \boldsymbol{U}_{\phi} , \boldsymbol{R}_{\delta}\right\}$
as the overall learnable weights in the GNN of client $k$. 
Consequently, we formulate \ours as a distributed optimization problem as follows:
\begin{equation}
\small{\min_{\boldsymbol{W}} F(\boldsymbol{W}) \stackrel{\text { def }}{=} \min_{\boldsymbol{W}} \sum_{k=1}^{K} \frac{N^{(k)}}{N} \cdot f^{(k)}(\boldsymbol{W}), \label{eq:FL}}
\end{equation}
where $\space f^{(k)}(\boldsymbol{W}) = \frac{1}{N^{(k)}} \sum_{i=1}^{N^{(k)}} \mathcal{L}(\boldsymbol{W}; x_{i}^{(k)}, z_{i}^{(k)}, y_{i}^{(k)})$ is the $k^{th}$ client's local objective function that measures the local empirical risk over the graph dataset $\mathcal{D}^{(k)}$ with $N^{(k)}$ data samples. $\mathcal{L}$ is the loss function of the global GNN model. To solve this problem, the most straightforward algorithm is FedAvg \citep{mcmahan2017communication}. It is important to note here that in FedAvg, the aggregation function on the server merely averages model parameters. We use GNNs inductively (i.e., the model is independent of the structure of the graphs it is trained on). Thus, no topological information about graphs on any client is required on the server during parameter aggregation. Other advanced algorithms such as FedOPT \citep{reddi2020adaptive}, FedGKT \citep{he2020group}, and Decentralized FL \citep{he2019central, he2021spreadgnn} can also be applied. 

Under the unified framework of \ours, we organize various distributed graph scenarios motivated by real-world applications into three settings based on how the graphs are distributed across silos, and provide support to the corresponding typical tasks in each setting.  
\begin{myitemize}
    \item \textbf{Graph-level \ours:} Each client holds a set of graphs, where the typical task is graph classification/regression. Real-world scenarios include molecular trials \citep{rong2020self}, protein discovery \citep{yang2018learned} and so on, where each institute might hold a limited set of graphs with ground-truth labels due to expensive experiments.
    \item \textbf{Subgraph-level \ours:} Each client holds a subgraph of a larger global graph, where the typical task is node classification and link prediction. Real-world scenarios include recommendation systems \citep{yang2021consisrec}, knowledge graph completion \citep{chen2020survey} and so forth, where each institute might hold a subset of user-item interaction data or entity/relation data.
    \item \textbf{Node-level \ours:} Each client holds the ego-networks of one or multiple nodes, where the typical task is node classification and link prediction. Real-world scenarios include social networks \citep{zhou2008brief}, sensor networks \citep{yick2008wireless}, etc., where each node only sees its $k$-hop neighbors and their connections in the large graph. 
\end{myitemize}

\paragraph{Supported GNN models and FL algorithms.}  \texttt{FedGraphNN}'s latest release supports the GNN models of GCN \citep{kipfgcn}, GAT \citep{gat}, GraphSage \citep{graphsage}, SGC \citep{wu2019simplifying}, and GIN \citep{xu2019powerful}, implemented via PyTorch Geometric \citep{fey2019fast}. For FL algorithms, aside from FedAvg \citep{mcmahan2017communication}, other advanced algorithms such as FedOPT \citep{reddi2020adaptive} are also supported within FedML library \cite{chaoyanghe2020fedml}. We refer to the Appendix \ref{app:baseline} for more details on the supported GNN baselines.


\vspace{-5pt}
\section{\ours Open Datasets}
\label{sec:datasets}
\vspace{-5pt}




\ours is centered around three federated GNN settings based on the ways graph data can be distributed in real-world scenarios, which covers a broad range of domains, tasks and challenges of graph FL.
Specifically, it includes 36 datasets from 7 domains, such as molecules, proteins, knowledge graphs, recommendation systems, citation networks and social networks.
Here, to facilitate clear understanding over the various graph FL settings, we organize and introduce examples of real-world datasets in each of the three federated GNN settings.
Exact sources and statistics of the datasets are provided in Table \ref{tab:dataset}, while more details are provided in Appendix \ref{app:datasets}. 

\begin{myitemize}
    \item \textbf{Graph-level Setting:} 
    In the real world, biomedical institutions might hold their own set of graphs such as molecules and proteins, and social network companies might hold their own set of community graphs. Such graphs may constitute large and diverse datasets for GNN traning, but they cannot be directly shared across silos. To simulate such scenarios, we utilize datasets from the domains of molecular machine learning  \cite{wu2018moleculenet}, bioinformatics \cite{protein-kernel,ddi,graphsage} and social computing \cite{dgk},  We also introduce a new large-scale dataset, called \texttt{hERG} \citep{chembl2017} for federated drug discovery.  
    \item \textbf{Subgraph-level Setting:} The first realistic scenario of subgraph-level FL is recommendation systems, where the users can interact with items owned by different shops or sectors, which makes each data owner only holding a part of the global user-item graph. To simulate such scenarios, we use recommendation datasets from both publicly available sources \citep{ciao, epinions} and internal sources \citep{Tencent}, which have high-quality meta-data information. 
    Another realistic scenario is knowledge graphs, where different organizations or departments might only have a subset of the entire knowledge, due to the focus in particular domains. We integrate the FB15k-237 \citep{fb15k-237}, WN18RR \citep{wn18rr} and YAGO3-10 \citep{yago3-10} datasets, where subgraphs can be build based on relation types to distinguish specialized fields or communities to distinguish the entities of focus.
    \item \textbf{Node-level Setting:} In social networks, each user's personal data can be sensitive and only visible to his/her $k$-hop neighbors (e.g., in Instagram, $k=1$ for contents and $k=2$ for links, of private accounts). Thus, it is natural to consider node-level FL in social networks with clients holding the user ego-networks. To simulate this scenario, we use the open social networks \citep{coauthor} and publication networks \citep{cora, corafull, citeseer, pubmed, dblp} and partition them into sets of ego-networks.
\end{myitemize}

\begin{table}[t]
  \centering
  \caption{Summary of open graph datasets from various domains contained in \ours.}
  \scalebox{0.7}{
     \begin{tabular}{rrlrrrrr}
     \toprule
    \multicolumn{1}{l}{\textbf{Task-Level}} & \multicolumn{1}{l}{\textbf{Category}}  & \textbf{Datasets} & \multicolumn{1}{l}{\textbf{\# Graphs}} & \multicolumn{1}{l}{\textbf{Avg. \# Nodes}} & \multicolumn{1}{l}{\textbf{Avg. \# Edges}} & \multicolumn{1}{l}{\textbf{Avg. Degree}} & \multicolumn{1}{l}{\textbf{\# Classes}}  \\
    \midrule
          \multicolumn{1}{l}{Graph-Level}& \multicolumn{1}{l}{Molecules} & BACE\citep{subramanian2016computational} & 1513  & 34.12 & 36.89 & 2.16 & 2      \\
          &       &        HIV\cite{HIV}   & 41127 & 25.53 & 27.48 & 2.15 & 2      \\
          &       &        MUV\cite{MUV}   & 93087 & 24.23 & 26.28 & 2.17  &  17  \\
          &       &        Clintox \citep{gayvert2016data} & 1478  & 26.13 & 27.86 & 2.13 & 2      \\
          &       &        SIDER \citep{kuhn2016sider} & 1427  & 33.64 & 35.36 & 2.10 &  27  \\
          &       &        Toxcast\cite{richard2016toxcast} & 8575  & 18.78 & 19.26 & 2.05 &  167  \\
          &       &        Tox21 \citep{tox21} & 7831  & 18.51 & 25.94 & 2.80 &  12  \\
          &       &        BBBP \citep{martins2012bayesian} & 2039  & 24.05 & 25.94 & 2.16 & 2      \\
          &       &        QM9 \citep{gaulton2012chembl}  & 133885 & 8.8   & 27.6  & 6.27 & 1      \\
          &       &        ESOL \citep{delaney2004esol}  & 1128  & 13.29 & 40.65 & 6.11 & 1      \\
          &       &        FreeSolv\citep{mobley2014freesolv}  & 642   & 8.72  & 25.6  &  5.87 & 1      \\
          &       &        Lipophilicity\citep{gaulton2012chembl}  & 4200  & 27.04 & 86.04 & 6.36 & 1      \\
          &       &        hERG \citep{chembl2017} & 10572 & 29.39 & 94.09 & 6.40 & 1      \\
          &       & MUTAG\citep{mutag} & 188   & 17.93 & 19.79 & 2.21 & 2       \\
          &       &        NCI1\citep{NCI1}& 4110  & 29.87 & 32.3  & 2.16 & 2       \\
            \cdashline{2-8}
          & \multicolumn{1}{l}{Proteins}  & PROTEINS\citep{protein-kernel}  & 1113  & 39.06 & 72.82 & 3.73 & 2       \\
          &       &        DDI \cite{ddi}   & 1178  & 284.32 & 715.66 & 5.03 & 2       \\
          &       &        PPI\cite{graphsage}   & 24 & 56,944 & 818,716 & 28.76 &  121   \\
          \cdashline{2-8}
           & \multicolumn{1}{l}{Social networks}  & COLLAB\citep{dgk} & 5000  & 74.49 & 2457.78 & 65.99 & 3       \\
          &       &        REDDIT-B\citep{dgk} & 2000  & 429.63 & 497.75 & 2.32 & 2       \\
          &       &        REDDIT-M-5K\citep{dgk} & 4999  & 508.52 & 594.87 & 2.34 & 5       \\
          &       &        IMDB-B\citep{dgk} & 1000  & 19.77 & 96.53 & 9.77 & 2       \\
          &       &        IMDB-M\citep{dgk} & 1500  & 13    & 65.94 & 10.14 & 3       \\
    \midrule
    \multicolumn{1}{l}{Subgraph-Level} 
          & \multicolumn{1}{l}{Recomm.~systems} & Ciao \citep{ciao}  & 28     & 5150.93 & 19280.93 & 3.74 &   5     \\
          &        & Epinions \citep{epinions} & 27    & 15824.22 & 66420.52 & 4.20 & 5      \\
          &        & Tencent \citep{Tencent}   & 1 & 709074 & 991713 & 2.80 & 2 \\
          \cdashline{2-8}
          & \multicolumn{1}{l}{Knowledge graphs} & FB15k-237 \citep{fb15k-237} & 1     & 14505 & 212110   & 14.62 & 237      \\
          &       &        WN18RR \citep{wn18rr} & 1     & 40559 &  71839  &  1.77  & 11       \\
          &       &        YAGO3-10 \citep{yago3-10} & 1     & 123143 &  774182  &  6.29  & 37       \\
    \midrule
    \multicolumn{1}{l}{Node-level} & \multicolumn{1}{l}{Publication networks} 
        & CORA \citep{cora} & 1 & 2708 & 5429 & 2.00 & 7 \\
        & & CORA-full \citep{corafull}  & 1 & 19793  & 65311  & 3.30 & 70       \\
        & & CITESEER \citep{citeseer} & 1     & 4230  & 5358  & 1.27 & 6       \\
        & & PUBMED \citep{pubmed} & 1     & 19717 & 44338 & 2.25 & 3       \\
        & & DBLP \citep{dblp} & 1 & 17716 & 105734 & 5.97 & 4 \\
        \cdashline{2-8}
    & \multicolumn{1}{l}{Social networks} & CS \citep{coauthor} & 1 & 18333 & 81894 & 4.47 & 15 \\
    & & Physics \citep{coauthor} & 1 & 34493 & 247962 & 7.19 & 5 \\
    \bottomrule
    \end{tabular}%
  }
  \label{tab:dataset}%
\end{table}%

In terms of graph mining tasks, \ours supports all three common tasks of graph classification, node classification and link prediction. Some tasks are naturally important in certain graph FL settings while others are not, which we also clarify with real examples as follows
\begin{myitemize}
    \item \textbf{Graph Classification:} This task is to categorize different types of graphs based on their structure and overall information. Unlike other tasks, this requires to characterize the property of the entire input graph. 
    This task is naturally important in graph-level FL, with real examples such as molecule property prediction, protein function prediction, and social community classification.
    \item \textbf{Link Prediction:} This task is to estimate the probability of links between any two nodes in a graph. 
    It is important in the subgraph-level FL, for example, in recommendation systems and knowledge graphs, where link probabilities are predicted in the former, and relation types are predicted in the latter.  It is less likely but still viable in the node-level setting, where friend suggestion and social relation profiling can be attempted in users' ego-networks, for example. 
    \item \textbf{Node Classification:} This task is to predict the labels of individual nodes in graphs. 
    It is more important in node-level FL, such as predicting the active research fields of an author based on his/her $k$-hop collaborators or habits of a user based on his/her $k$-hop friends. It might also be important in subgraph-level FL, such as the collaborative prediction of disease infections based on the patient networks dispersed in multiple healthcare facilities. 
\end{myitemize}

\paragraph{Data sources.} We exhaustively examine and collecte 36 datasets from 7 domains. Among them, 34 are from publicly available sources such as MoleculeNet \cite{wu2018moleculenet} and graph kernels datasets \cite{protein-kernel}. In addition, we introduce two new de-identified datasets based on our collaboration with Tencent: \textit{hERG} \citep{kim2021pubchem,gaulton2017chembl}, a graph dataset for classifying protein molecules responsible for cardiac toxicity and \textit{Tencent} \cite{He2019CascadeBGNNTE}, a large bipartite graph representing the relationships between users and groups. More details and their specific preprocessing procedures can be found in Appendices \ref{app:datasets:sources} \& \ref{app:datasets:pp}.  We plan to continually enrich the available datasets in the future through active collection of open datasets and collaboration with industrial partners.
\paragraph{Generating Federated Learning Datasets .} Unlike traditional ML banchmarking datasets, graph datasets and real-world graphs may exhibit non-I.I.D.-ness due to sources such as structure and feature heterogeneity \cite{yang2020heterogeneous,yang2020co,yang2019conditional}. Coupled with FL, multiple sources of non-I.I.D-ness are indistinguishable.  Here, our focus is on how to produce \textit{reproducible} and \textit{statistical (sample-based)} non-I.I.D.ness. To generate \textit{sample-based} non-I.I.D.ness, we use the unbalanced partition algorithm Latent Dirichlet Allocation (LDA) \citep{chaoyanghe2020fedml} to partition datasets, and this method can be applied regardless of data domain. In addition, researchers and practitioners can also synthesize non-I.I.D.-ness using our available additional meta-data. Figure \ref{fig:noniid_dists} in the Appendix shows several datasets' non-I.I.D. distributions generated using the LDA method. The alpha values for LDA for representative datasets can be found in Tables \ref{tab:graphlevel},\ref{tab:subgraphlevel},\ref{tab:nodelevel}
 and \ref{tab:graphlevel-reg} in the Appendix \ref{app:experiments:more}. Yet, deeply decoupling and quantifying non-I.I.D.-ness in federated GNNs remains as an open problem \cite{wang2020noniid, xie2021federated}.

In summary, we provide a comprehensive study and solutions for several challenges in data collection and benchmark for graph FL: (1) Collecting, analyzing and categrozing a large number of public, real-world datasets into different federated GNN settings with corresponding typical tasks; (2) Standardizing the procedure to synthesize non-I.I.D.~data distributions for all graph-structured datasets through providing a novel partition method and associative meta-data.

\vspace{-5pt}
\section{\ours Benchmark System: Efficient, Secure, and Modularized}
\label{sec:design}
\vspace{-5pt}

We have released an open-source distributed training system for \texttt{FedGraphNN}. This system is tailored for benchmarking graph FL and promoting algorithmic innovations with three key advantageous designs in the context of FL that have not been supported by existing simulation-oriented benchmark and libraries \cite{chaoyanghe2020fedml}.

\begin{figure}[htb!]
\centering
{\includegraphics[scale=0.21]{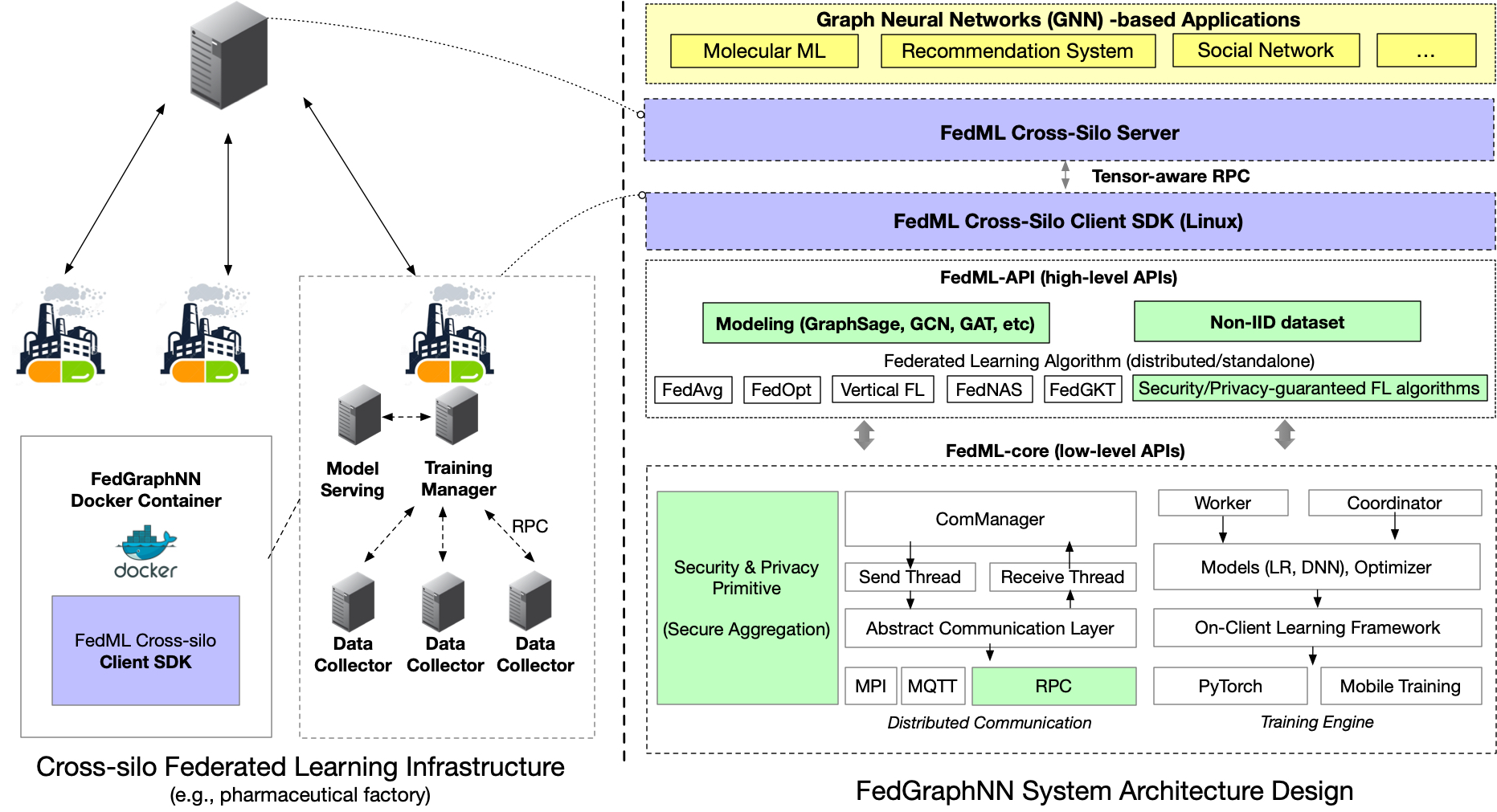}}
\caption{Overview of \texttt{FedGraphNN} System Architecture Design}
\label{fig:architecture}
\end{figure}

\paragraph{Enhancing realistic evaluation with efficient and deployable distributed system design.}

We design the training system to support realistic distributed computing in multiple edge servers, given that \ours is mainly executed in the cross-silo settings where each FL client represents an edge server belonging to an organization rather than smartphone or IoT devices. The system architecture, shown in Figure~\ref{fig:architecture}, is composed of three layers: \texttt{FedML-core} layer, \texttt{FedML-API} layer, and \texttt{Application} layer. \texttt{FedML-core} layer supports both RPC (remote procedure call) and MPI (message passing interface), which enable communication among edge servers located at different data centers. More specially, the RPC API is tensor-oriented, meaning that its weight or gradient transmit among clients is much faster than na\"{i}ve gRPC with GPU-direct communication (e.g., data owners from different AWS EC2 accounts can utilize this feature for faster training). The communication primitives are wrapped as abstract communication APIs (i.e., \texttt{ComManager} in Figure~\ref{fig:architecture}) to simplify the message definition and passing requested by different FL algorithms in \texttt{FedML-API} layer (see more details in Appendix \ref{app:system}). In the deployment perspective, the FL client library should be compatible with heterogeneous hardware and OS configuration. To meet this goal, we provide Docker containers to simplify the large-scale deployment for federated training.

With the help of the above design, researchers can run realistic evaluations in a parallel computing environment where multiple CPU/GPU servers are located in multiple organizations (e.g., edge servers in AWS EC2). As such, for medium and small-scale graph datasets, the training can be finished in only a few minutes. For large-scale graph datasets, researchers can also measure system-wise performance (communicational and computational costs) to have a tangible trade-off between accuracy and system efficiency. Scaling up to numerous edge servers (FL clients) is further simplified by Docker-based deployment.

\paragraph{Enabling secure benchmarking with lightweight secure aggregation.} Researchers in the industry may also need to explore FL on their private customer datasets with other organizations. However, model weights from clients may still have the risk of privacy leakage \cite{zhu2020deep}. As such, legal and regulatory departments normally do not permit FL research on private customer datasets when strong security is not guaranteed. To break this barrier, we integrate secure aggregation (SA) algorithms to the \ours system. ML researchers do not need to master security-related knowledge but also enjoy a secure distributed training environment. To be more specific, we support \ours with \texttt{LightSecAgg}, a state-of-the-art SA algorithm developed by our team (Appendix \ref{app:lightsecagg}). In high-level understanding, \texttt{LightSecAgg} protects the client model by generating a single random mask and allows their cancellation when aggregated at the server. Consequently, the server can only see the aggregated model and not the raw model from each client. 
The design and implementation of \texttt{LightSecAgg} spans across \texttt{FedML-core} and \texttt{FedML-API} in the system architecture, as shown in Figure~\ref{fig:architecture}. Baselines such as SecAgg \cite{bonawitz2017practical} and SecAgg+ \cite{bell2020secure} are also supported.

\begin{figure}[h!]
\centering
{\includegraphics[scale = 0.33]{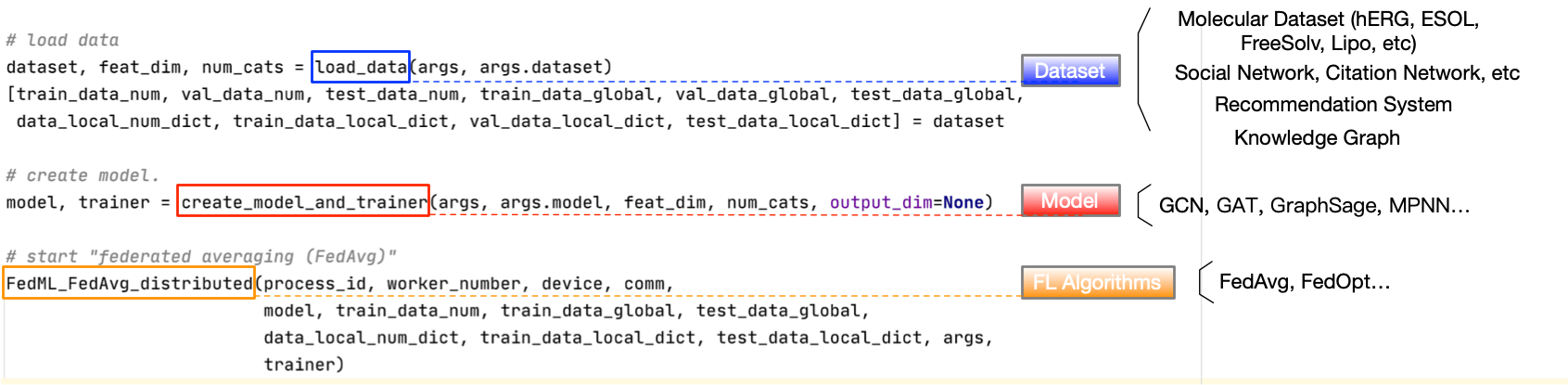}}
\caption{Example code for benchmark evaluation with \texttt{FedGraphNN}}
\label{fig:code}
\end{figure}

\paragraph{Facilitating algorithmic innovations with diverse datasets, GNN models, and FL algorithms.}\texttt{FedGraphNN} also aims to enable flexible customization for future algorithmic innovations. 
To support diverse datasets, GNN models, and FL algorithms, we have modularized the API and component design. All data loaders follow the same format of input and output arguments, which are compatible with different models and algorithms, and easy to support new datasets. The method of defining the model and related trainer is kept the same as in centralized training to reduce the difficulty of developing the distributed training framework. For new FL algorithm development, worker-oriented programming reduces the difficulty of message passing and definition (details are introduced in the Appendix \ref{app:system}). Diverse algorithmic implementations serve as the reference source code for future algorithmic innovation. The user-oriented interface (main training script) is simplified as the example code shown in Figure \ref{fig:code} where a few lines of code can launch a federated training in a cross-silo cloud environment.

\vspace{-5pt}
\section{\ours Empirical Analysis}
\label{sec:experiments}
\vspace{-5pt}


\subsection{Experimental Setup}
\vspace{-5pt}

Our experiments are conducted on multiple GPU servers each equipped with 8 NVIDIA Quadro RTX 5000 (16GB GPU memory). The hyper-parameters are selected via our built-in efficient parameter sweeping functionalities from the ranges listed in Appendix \ref{app:experiments:hp}. We present results on the ROC-AUC metric for graph classification and RMSE \& MAE for graph regression, MAE, MSE, and RMSE for link prediction, and micro-F1 for node classification. More evaluation metrics supported by \ours are presented in Appendix \ref{app:experiments:metric}.


\subsection{Baseline Performance Analysis}

We report experimental results of several popular GNN models trained with the most widely used FL algorithm of FedAvg, to examplify the utility of \ours. More results with varying baselines, hyper-parameters, evaluation metrics and visualizations are being updated and presented in Appendix \ref{app:experiments:more}. 
After hyper-parameter tuning, we present the main performance results as well as runtimes in Tables \ref{tab:graphlevel}, \ref{tab:subgraphlevel} and \ref{tab:nodelevel}. 

\begin{table*}[htbp]
\centering
\caption{Performance of graph classification in the graph-level FL setting (\#clients=4).}
\resizebox{\textwidth}{!}{
    \begin{tabular}{l  c c c c c c c c c c  }
    \toprule
     Metric & \multicolumn{5}{c}{ROC-AUC} & \multicolumn{5}{c}{Training Time (sec.)} \\ 
     \cmidrule(lr){2-6}\cmidrule(lr){7-11}
     Method & \textsc{SIDER } & \textsc{BACE} & \textsc{Clintox} & \textsc{BBBP} & \textsc{Tox21} & \textsc{SIDER} & \textsc{BACE} & \textsc{Clintox} & \textsc{BBBP} & \textsc{Tox21} \\
      & \textsc{$\alpha = 0.2$} & \textsc{$\alpha = 0.5$} & \textsc{$\alpha = 0.5$} & \textsc{ $\alpha = 2$} & \textsc{$\alpha = 3$} &  & & &  &  \\
     \midrule
        MoleculeNet Results & 0.6380 & 0.8060 & 0.8320 & 0.6900 & 0.8290  & \multicolumn{5}{c}{Not Published} \\
      \midrule
        GCN (Centralized) & 0.6476 & 0.7657 & 0.8914 & 0.8705 & 0.7800  & 458 & 545 & 686 & 532 & 1034  \\
     GCN (FedAvg) & 0.6266 & 0.6594 & 0.8784 & 0.7629 & 0.7128 & 358 & 297 & 280 & 253 & 903 \\ 
         \midrule
        GAT (Centralized) & 0.6639 & 0.9221 & 0.9573 & 0.8824 & 0.8144 & 739 & 603 & 678 & 533 & 2045  \\
     GAT (FedAvg) & 0.6591 & 0.7714 & 0.9129 & 0.8746 & 0.7186 & 528 & 327 & 457 & 328 & 1549 \\ 
       \midrule
        GraphSAGE (Centralized) & 0.6669 & 0.9266 & 0.9716 &0.8930 & 0.8317 & 193 & 327 & 403 & 312 & 1132 \\
     GraphSAGE (FedAvg) & 0.6700 & 0.8604 & 0.9246 & 0.8935 & 0.7801 & 127 & 238 & 282 & 206 & 771   \\ 
    \bottomrule
    \end{tabular}
}
\label{tab:graphlevel}
\end{table*}

\begin{table}[htbp]
\centering
\caption{Performance of link prediction in the subgraph-level FL setting (\#clients = 8).}
\resizebox{\textwidth}{!}{
    \begin{tabular}{lcccccccc}
    \toprule
       Metric & \multicolumn{2}{c}{MAE} & \multicolumn{2}{c}{MSE} & \multicolumn{2}{c}{RMSE} & \multicolumn{2}{c}{Training Time (sec.)} \\ 
      \cmidrule(lr){2-3}\cmidrule(lr){4-5}\cmidrule(lr){6-7}\cmidrule(lr){8-9}
      DataSet &  \textsc{Ciao} & \textsc{Epinions} &  \textsc{Ciao} & \textsc{Epinions} &  \textsc{Ciao} & \textsc{Epinions} &  \textsc{Ciao} & \textsc{Epinions}\\
      \midrule
        GCN (Centralized) & 0.8167 &  0.8847 & 1.1184 &1.3733  & 1.0575 & 1.1718 & 268 & 650 \\
        GCN (FedAvg) & \colorbox{aurometalsaurus!20}{0.7995} & 0.9033  & \colorbox{aurometalsaurus!20}{1.0667} & 1.4378 & \colorbox{aurometalsaurus!20}{1.0293} & 1.1924  & 352 & 717  \\
      \midrule
        GAT (Centralized) & 0.8214  & 0.8934  & 1.1318  & 1.3873  & 1.0639  & 1.1767 & 329 & 720  \\
         GAT (FedAvg) & \colorbox{aurometalsaurus!20}{0.7987} & 0.9032  &  \colorbox{aurometalsaurus!20}{1.0682} & 1.4248 & \colorbox{aurometalsaurus!20}{1.0311} & 1.1882 & 350 & 749 \\
      \midrule
        GraphSAGE (Centralized) & 0.8231  & 1.0436  & 1.1541 & 1.8454 & 1.0742 & 1.3554 & 353 & 721 \\
        GraphSAGE (FedAvg) & 0.8290 & \colorbox{aurometalsaurus!20}{0.9816} & \colorbox{aurometalsaurus!20}{1.1320} & \colorbox{aurometalsaurus!20}{1.6136} & \colorbox{aurometalsaurus!20}{1.0626}  & \colorbox{aurometalsaurus!20}{1.2625}  & 551 & 810 \\
    \bottomrule
    \end{tabular}
}
\label{tab:subgraphlevel}
\end{table}


\begin{table*}[htbp]
\centering
\caption{Performance of Node classification in the node-level FL setting (\#clients = 10).}
\resizebox{\textwidth}{!}{
    \begin{tabular}{l l c c c c c c c }
    \toprule
      Metric & \multicolumn{4}{c}{micro F1} & \multicolumn{4}{c}{Training Time (sec.)} \\ 
     \cmidrule(lr){2-5}\cmidrule(lr){6-9}
     Method & \colorbox{white!20}{\textsc{CORA}} & \textsc{CITESEER} & \textsc{PUBMED} & \textsc{DBLP} & \textsc{CORA} & \textsc{CITESEER} & \textsc{PUBMED} & \textsc{DBLP} \\
      \midrule
      GCN (Centralized) & \colorbox{white!20}{0.8622} &  0.9820 &  0.9268 &  0.9294 & 1456 & 742 & 1071 & 1116\\
      GCN (FedAvg) & \colorbox{white!20}{0.8549} & 0.9743 & 0.9128 & 0.9088  & 833 & 622 & 654 & 653\\
      \midrule
      GAT (Centralized)  & \multirow{2}{*}{\colorbox{white!20}{diverge}} & 0.9653 & 0.8621 & 0.8308 & 1206 & 1765 & 1305 & 957\\
      GAT (FedAvg)    &  & 0.9610 & 0.8557 & 0.8201 & 871 & 652 & 682 & 712 \\
      \midrule
      GraphSAGE (Centralized)  & \colorbox{white!20}{0.9692} & 0.9897 & 0.9724 & 0.9798 & 1348 & 934 & 692 & 993\\
      GraphSAGE (FedAvg) & \colorbox{aurometalsaurus!20}{0.9749} & 0.9854 & \colorbox{aurometalsaurus!20}{0.9761} & 0.9749 & 774 & 562 & 622 & 592\\
    \bottomrule
    \end{tabular}
}
\label{tab:nodelevel}
\end{table*}

Besides showcasing the utility of \ours, there are multiple takeaways from these results:

\begin{enumerate}[leftmargin=0.5cm]
    \item When the graph datasets are small, FL accuracy is often on par with centralized learning.
    \item When dataset sizes grow, FL accuracy becomes significantly worse than centralized learning. We conjecture that it is the complicated non-I.I.D.~nature of graphs that leads to the accuracy drop in larger datasets.
    \item The dynamics of training GNNs in a federated setting are different from training federated vision or language models. Our findings show that the best model in the centralized setting may not necessarily be the best model in the FL setting. 
    \item Counterintuitive phenomenons (highlights in above tables) further add to the mystery of federated graph neural networks: in graph-level experiments, GAT suffers the most performance compromise on 5 out of 9 datasets; in both subgraph-level and node-level FL, results on some datasets (CIAO, CORA, PubMed) may even have slightly higher performance than centralized training; GAT cannot achieve reasonable accuracy in node-level FL (e.g., in CORA dataset), etc.
\end{enumerate} 
These results indicate the limitations of the baselines in \ours and motivate much further research in understanding the nuances and improving the training GNNs in the FL setting.


\paragraph{Evaluation on System Efficiency and Security.} We provide additional results on system performance evaluation, where the results are summarized in Appendix \ref{app:system-analysis}. Depending on the size of the graph data, \ours can complete the training efficiently. The training time ranges from a few minutes to about 1 hour even for large-scale graphs. In the security aspect, the main result of \texttt{LightSecAgg} is provided in Appendix \ref{app:lightsecagg}. The key benefit is that it not only obtains the same level of privacy guarantees as to the state-of-the-art (SecAgg [7] and SecAgg+ [3]) but also substantially reduces the aggregation complexity (hence much faster training).


\vspace{-5pt}
\section{Related Works and Open Challenges}
\label{sec:related_works} 
\vspace{-10pt}


\ours lies at the intersection of GNNs and FL. We first discuss related works under the umbrella of three different graph FL settings. (1) \textit{Graph-level} (Figure \ref{fig:sub-first}): we believe molecular machine learning is a paramount application in this setting, where many small graphs are distributed between multiple institutions, as demonstrated in \citep{he2021spreadgnn,xie2021federated}. \cite{xie2021federated} proposes a clustered FL framework specifically for GNNs to deal with feature and structure heterogeneity. \cite{he2021spreadgnn} develops a multi-task learning framework suitable to train federated graph-level GNNs without the need for a central server.
(2) \textit{Subgraph-level} (Figure \ref{fig:sub-second}): this scenario typically pertains to the entire social networks, recommender networks or knowledge graphs that need to be partitioned into many smaller subgraphs due to data barriers between different departments in a giant company or data platforms with different domain focuses as demonstrated in \citep{wu2021fedgnn, zhang2021subgraph}. \citep{wu2021fedgnn} proposes a federated recommendation system with GNNs, whereas \cite{zhang2021subgraph} proposes FedSage, a subgraph-level federated GNN generating psuedo-neighbors utilizing variational graph autoencoder. (3) \textit{Node-level} (Figure \ref{fig:sub-third}): when the privacy of specific nodes in a graph is important, node-level graph FL is useful in practice. The IoT setting is a good example \cite{zheng2020asfgnn};  \citep{wang2020graphfl} uses a hybrid method of FL and meta-learning to solve the semi-supervised graph node classification problem in decentralized social network datasets; \citep{meng2021crossnode} attempts to protect the node-level privacy using an edge-cloud partitioned GNN model for spatio-temporal forecasting tasks using node-level traffic sensor datasets.

Before our unified system of \ours, there was a serious lack of standardized datasets and baselines for training graph-neural networks in a federated setting pertains. Previous platforms like LEAF \cite{caldas2019leaf}, TensorFlow Federated \cite{tensorflow2015-whitepaper,tff}, PySyft \cite{ryffel2018generic}, and FATE have no support on graph datasets and GNN models. 
Beyond the direct goals of \ours, many open algorithmic challenges in graph FL remain be studied. First, integrating both graph topology of GNNs and network topology of FL in a principled and efficient way is highly dataset- and application-specific. Second, the partitioning of a single graph into subgraphs or ego-networks for some tasks introduce dataset bias and information loss in terms of missing cross-subgraph links \cite{zhang2021subgraph}. Third, decoupling and quantifying the multiple sources of non-I.I.D.-ness coming from both features and structures in graphs are crucial for the appropriate design of federated GNNs \cite{xie2021federated,wang2020noniid}. Finally, in terms of communication efficiency and security, previous works utilize various methods including Secure Multi-Party Computation (SMPC) \citep{zhou2020privacy}, Homomorphic Encryption (HE) \citep{zhou2020privacy},  secure aggregation \citep{jiang2020federated} and Shamir's secret sharing \citep{zheng2020asfgnn}, but their comparison is missing.

\vspace{-10pt}
\section{Conclusions and Future Works} 
\label{sec:conclusion} 
\vspace{-10pt}

In this work, we design an FL system and benchmark for GNNs, named \ours, which includes open datasets, baseline implementations, programmable APIs, all integrated in a robust system affordable to most research labs. We hope \ours can serve as an easy-to-use research platform for researchers to explore vital problems at the intersection of FL and GNNs. 

Here we highlight some future improvements and research directions based on our \ours system: 1. supporting more graph datasets and GNN models for diverse applications. Possible applications include and are not limited to sensor networks and spatio-temporal forecasting \citep{liu2021dig,yang2018did}; 2. optimizing the system to further accelerate the training speed for large graphs \citep{zheng2020distdgl,9378178}; 3. designing advanced graph FL algorithms to mitigate the accuracy gap on datasets with non-I.I.D.ness, such as tailoring FedNAS \citep{he2020fednas,he2020milenas} to search for personalized GNN models for individual FL clients; 4. exploring label-efficient GNN models based on concepts such as meta-learning and self-supervision to exploit the graphs in each client and their collaboration \citep{xie2021self}; 5. addressing challenges in security and privacy under the setting of Federated GNN \citep{elkordy2020secure,prakash2020mitigating,prakash2020coded1,prakash2020coded2,yang2021secure,chen2021understanding}; 6. proposing efficient compression algorithms that adapt to the level of compression to the available bandwidth of the users while preserving the privacy of users' local data; 7. organizing data competitions, themed workshops, special issues, etc., on the dissemination of \ours; 8. actively discussing ethics and societal impacts to avoid unwanted negative effects. 







\bibliography{iclr2021_conference}
\bibliographystyle{iclr2021_conference}

\clearpage
\section*{Checklist}


\begin{enumerate}

\item For all authors...
\begin{enumerate}
  \item Do the main claims made in the abstract and introduction accurately reflect the paper's contributions and scope?
    \answerYes{}
  \item Did you describe the limitations of your work?
    \answerYes{}
  \item Did you discuss any potential negative societal impacts of your work?
    \answerYes{}
  \item Have you read the ethics review guidelines and ensured that your paper conforms to them?
    \answerYes{}
\end{enumerate}

\item If you are including theoretical results...
\begin{enumerate}
  \item Did you state the full set of assumptions of all theoretical results?
    \answerNA{}
	\item Did you include complete proofs of all theoretical results?
    \answerNA{}
\end{enumerate}

\item If you ran experiments (e.g. for benchmarks)...
\begin{enumerate}
  \item Did you include the code, data, and instructions needed to reproduce the main experimental results (either in the supplemental material or as a URL)?
    \answerYes{}
  \item Did you specify all the training details (e.g., data splits, hyperparameters, how they were chosen)?
    \answerYes{}
	\item Did you report error bars (e.g., with respect to the random seed after running experiments multiple times)?
    \answerNo{}
	\item Did you include the total amount of compute and the type of resources used (e.g., type of GPUs, internal cluster, or cloud provider)?
    \answerYes{}
\end{enumerate}

\item If you are using existing assets (e.g., code, data, models) or curating/releasing new assets...
\begin{enumerate}
  \item If your work uses existing assets, did you cite the creators?
    \answerYes{}
  \item Did you mention the license of the assets?
    \answerNo{}
  \item Did you include any new assets either in the supplemental material or as a URL?
    \answerNo{}
  \item Did you discuss whether and how consent was obtained from people whose data you're using/curating?
    \answerYes{}
  \item Did you discuss whether the data you are using/curating contains personally identifiable information or offensive content?
    \answerYes{}
\end{enumerate}

\item If you used crowdsourcing or conducted research with human subjects...
\begin{enumerate}
  \item Did you include the full text of instructions given to participants and screenshots, if applicable?
    \answerNA{}
  \item Did you describe any potential participant risks, with links to Institutional Review Board (IRB) approvals, if applicable?
    \answerNA{}
  \item Did you include the estimated hourly wage paid to participants and the total amount spent on participant compensation?
    \answerNA{}
\end{enumerate}

\end{enumerate}

\clearpage

\appendix 

\section{More Details of the Supported Graph Neural Network Architectures}
\label{app:baseline}

\begin{myitemize}
    \item \textbf{Graph Convolutional Networks} \citep{kipfgcn} is a GNN model which is a $1^{st}$ order approximation to spectral GNN models.  \cite{markowitz2021graph}
    \item \textbf{GraphSAGE} \citep{graphsage} is a general inductive GNN framework capable of generating node-level representations for unseen data. 
    \item \textbf{Graph Attention Networks} \citep{gat} is the first attention-based GNN model. Attention is computed in a message-passing fashion.
      \item \textbf{Simplifying Graph Convolutional Networks} \citep{wu2019simplifying} is the first attention-based GNN model. Attention is computed in a message-passing fashion.
       \item \textbf{Graph Isomorphism Networks} \citep{xu2019powerful} is the SotA method on graph classification showing that GNNs are strong at most a 1-Weisfeiler Lehman test.
\end{myitemize}

\section{More Details of the Open Datasets}
\label{app:datasets}

\subsection{Data Sources}
\label{app:datasets:sources}

The details of each dataset are listed below:
 
\paragraph{Datasets for Graph-level \ours }
\begin{myitemize}
    \item \texttt{BBBP} \citep{martins2012bayesian} involves records of whether a compound carries the permeability property of penetrating the blood-brain barrier.
    \item \texttt{SIDER} \citep{kuhn2016sider}, or Side Effect Resource, the dataset consists of marketed drugs with their adverse drug reactions. The available 
    \item \texttt{ClinTox} \citep{gayvert2016data} includes qualitative data of drugs both approved by the FDA and rejected due to the toxicity shown during clinical trials.
    \item \texttt{BACE} \citep{subramanian2016computational} is collected for recording compounds that could act as the inhibitors of human $\beta$-secretase 1 (BACE-1) in the past few years.
    \item \texttt{Tox21}\citep{tox21} is a dataset which records  the toxicity of compounds. 
    \item \texttt{hERG}\citep{kim2021pubchem,gaulton2017chembl} is a dataset that records the gene (KCNH2) that codes for a protein known as Kv11.1 responsible for its contribution to the electrical activity of the heart to help the coordination of the heart's beating.
    \item \texttt{QM9} \citep{ramakrishnan2014quantum} is a subset of GDB-13, which records the computed atomization energies of stable and synthetically accessible organic molecules, such as HOMO/LUMO, atomization energy, etc. It contains various molecular structures such as triple bonds, cycles, amide, and epoxy.
    \item \texttt{ESOL}  \citep{delaney2004esol} is a small dataset documenting the water solubility(log solubility in mols per litre) for common organic small molecules.
    \item \texttt{Lipophilicity} \citep{gaulton2012chembl}  which records the experimental results of octanol/water distribution coefficient for compounds.
    \item  \texttt{FreeSolv} \citep{mobley2014freesolv} contains the experimental results of hydration-free energy of small molecules in water.
    \item \texttt{MUTAG} \cite{mutag} is a collection of nitroaromatic molecules and the aim is to classify their mutagenicity on Salmonella typhimurium. Input graphs are used to represent chemical compounds, where vertices stand for atoms and are labeled by the atom type (represented by one-hot encoding), while edges between vertices represent bonds between the corresponding atoms. It includes 188 samples of chemical compounds with 7 discrete node labels.
    
    \item \texttt{PROTEINS}\cite{protein-kernel} is a dataset of protein molecules and goal is to classify whether a protein is an enzyme or not. Nodes represent the amino acids and two nodes are connected by an edge if they are less than 6 Angstroms apart.
    
    \item \texttt{NCI1}\cite{NCI1} is a dataset where each input graph represents a  chemical compound in which  each vertex is an atom of the molecule(encoded via 1-hot vector), and edges between vertices represent bonds between atoms. This dataset is used to detect whether a chemical is responsible for cell lung cancer or not. 
    
    \item \texttt{DDI}\cite{ddi} is  a drug-drug interactions as well as documents describing drug-drug interactions from the DrugBank database.
    
    \item \texttt{COLLAB}\cite{dgk} is a scientific collaboration dataset consisting of researchers' ego networks, the graph representation of a researcher and his/her collaborators' collaborations. These graphs have three possible labels to distinguish researchers' field: High Energy Physics, Condensed Matter Physics, and Astro Physics.
    
    \item\texttt{IMDB-B} \& \texttt{IMDB-M}\cite{dgk} are movie collaboration datasets consisting of ego-networks of 1,000 actors/actresses featured in IMDB. In each graph, nodes represent actors/actresses, and there is an edge between them if they appear in the same movie. The main difference is that the latter one has more than 2 categories.
    
    \item\texttt{REDDIT-B} \& \texttt{REDDIT-M-5K}\cite{dgk} are relational datasets of graphs describing online discussions on Reddit where nodes represent users, and there is an edge between them if at least one of them respond to the other’s comment. The only difference is that graphs in \texttt{REDDIT-B} labeled according to whether it belongs to a question/answer-based community or not. There are multiple categories inside \texttt{REDDIT-B}.
    
\end{myitemize}

\paragraph{Datasets for Subgraph-level \ours}

Our focus is for recommendation systems.  A node in the graph represent a user or an item while the edge weight represents the rating score from user to item. Items are also assigned to different categories based on their characteristics. Each item belongs to at least one category.

\begin{myitemize}
    \item \texttt{Ciao} \citep{ciao} dataset contains rating information of users given to items, and also contain item category information. 

    \item \texttt{Epinions} \citep{epinions} dataset is trust network dataset containing  profile,  ratings and  trust relations of a user as triplet for each user in the network. For each rating, it has the product name and its category, the rating score, the time point when the rating is created, and the helpfulness of this rating.
    
    \item \texttt{Tencent} \citep{Tencent} dataset is a large bipartite graph representing the relation between users and groups. An edge indicates the user belongs to the connected group. The data also contains all the node features and node labels. 
    
    \item \texttt{FB15k-237} \citep{fb15k-237} contains knowledge base relation triples and textual mentions of Freebase entity pairs. 
    
    \item \texttt{WN18RR} \citep{wn18rr} is a link prediction dataset created from WordNet containing  93,003 triplets with 11 different relations of 40,943 entries.
    
    \item \texttt{YAGO3-10} \citep{yago3-10} or known as \texttt{Yet Another Great Ontology 3-10}, is a subset of YAGO, well-known benchmark dataset for knowledge base completion. Contains entities related to at least ten different relations. Triplets describe atributes like citizenship, profession, salary.

\end{myitemize}

\paragraph{Datasets for Node-level \ours}

\begin{myitemize}
    \item \texttt{CORA} \citep{cora} dataset is a citation network formed of 2708 scientific publications with 5429 links classified into one of seven classes. Each publication in the dataset is described by a 0/1-valued word vector indicating the absence/presence of the corresponding word from the dictionary. The dictionary consists of 1433 unique words.
    
    \item \texttt{CORA-full} \citep{corafull} dataset is an additional version of \texttt{cora} \citep{cora} which is extracted from the original entire network. It consists of 19,793 scientific publications with 65,311 links classified into one of 70 classes. The node features are represented by one-hot vectors indicating the absence/presence of words.

    \item \texttt{CITESEER} \citep{citeseer} is a citation network formed from 3312 scientific publications with 4732 links classified into one of six classes. Each publication in the dataset is described by a 0/1-valued word vector indicating the absence/presence of the corresponding word from the dictionary. The dictionary consists of 3703 unique words.
        
    \item \texttt{PUBMED} \citep{pubmed} dataset consists of 19717 scientific publications with 44338 links from PubMed database pertaining to diabetes classified into one of three classes. Each publication in the dataset is described by a TF/IDF weighted word vector from a dictionary which consists of 500 unique words.

    \item \texttt{DBLP} \citep{dblp} is a citation network dataset where it is extracted from DBLP, ACM, MAG, and other sources. The first version contains 629,814 papers and 632,752 citations. Each paper is associated with abstract, authors, year, venue, and title. The data set can be used for clustering with network and side information, studying influence in the citation network, finding the most influential papers, topic modeling analysis, etc.
    
    \item \texttt{Coauthor-CS} \citep{coauthor} is a coauthor network focusing on the area of computer science, in which nodes (18,333) represent authors and links (81, 894) indicate that the connected authors present on the same paper. The features of nodes represent paper keywords for each author’s papers, and the labels of nodes (15) indicate the most active study area of authors.
    
    \item \texttt{Coauthor-Physics} \citep{coauthor} is a coauthor network focusing on the area of physics which has the same representations of nodes, links, and labels as \texttt{Coauthor-CS}. It consists of 34,493 nodes and 247,962 links, and nodes are classified into one of 5 classes.

\end{myitemize}

\subsection{Data Preprocessing}
\label{app:datasets:pp}

\paragraph{Dataset Splitting.} Before generating non-I.I.D.'ness for our datasets, we partition all datasets  such that 80$\%$ training, 10$\%$ validation, and 10$\%$ test. This ratio can be modified and as future work, we plan to support domain-specific splits such as  scaffold splitting \citep{bemis1996properties}  for molecular machine learning datasets.

\paragraph{Molecular Datasets}
The feature extraction process is in two steps:
\begin{enumerate}
    \item  Atom-level feature extraction and Molecule object construction using RDKit \citep{landrum2006rdkit}.
    \item Constructing graphs from molecule objects using NetworkX \citep{SciPyProceedings_11}.
\end{enumerate}
Atom features, shown in  Table \ref{tab:atomfea}, are the atom features we used exactly the same as in \citep{grover}. 
\begin{table}[htbp]
\caption{Atom features}
\begin{center}
\begin{tabular}{lll}
\toprule
\multicolumn{1}{c}{\bf Features}  &\multicolumn{1}{c}{\bf Size} &\multicolumn{1}{c}{\bf Description}
\\ \midrule 

atom type & 100  &  Representation of atom (e.g., C, N, O), by its atomic number\\
    formal charge  & 5  & An integer electronic charge assigned to atom  \\
    number of bonds & 6 & Number of bonds the atom is involved in \\
    chirality & 5 & Number of bonded hydrogen atoms\\
    number of H & 5 & Number of bonded hydrogen atoms\\
    atomic mass & 1 & Mass of the atom, divided by 100\\
    aromaticity & 1 & Whether this atom is part of an aromatic system\\
    hybridization & 5 &  SP, SP2, SP3, SP3D, or SP3D2\\
    \bottomrule
\end{tabular}
\label{tab:atomfea}
\end{center}

\end{table}

\paragraph{Recommendation Systems/Knowledge Graph Datasets} For recommendation systems, the data pre-processing step is partitioning the bipartite graph into subgraphs by item categories. Items belong to the same category and related users form a subgraph. Subgraphs are combined if one client holds more than one category. In knowledge graph, the pre-processing includes two steps. We first build subgraphs by relation types or node community, and then partition the data to multiple clients by a specific manner (uniform or non-I.I.D. partitioning with LDA).

\paragraph{Citation/Coauthor Datasets} The main data pre-processing step for citation/coauthor networks in the node-level setting is to sample central nodes (egos) and build the $k$-hop neighborhoods correspondingly ($k$-hop ego-networks). For each dataset with a big entire graph, we randomly sample a number of egos (e.g. 1000) and construct $k$ (e.g. $2$)-hop egonetworks for them. These ego-networks are then partitioned by a specific manner (e.g. LDA non-IID partition) and distributed to multiple clients.

\subsection{Non-I.I.D. Partitioning}
\label{app:datasets:non-iid}

\paragraph{Latent Dirichlet Allocation(LDA)-Based}
According to \citep{yurochkin2019bayesian,wang2020federated}, we generate a heterogeneous partition into J clients by sampling $ p_{k} \sim \operatorname{Dir}_{J}(\alpha)$
and allocating a $p_{k,j}$ proportion of the training instances of class k to local client.

\begin{figure}[h!]
\centering
\subfigure[\label{fig:alpha0.1} $\alpha = 0.1$ ] {{\includegraphics[width=0.4\textwidth]{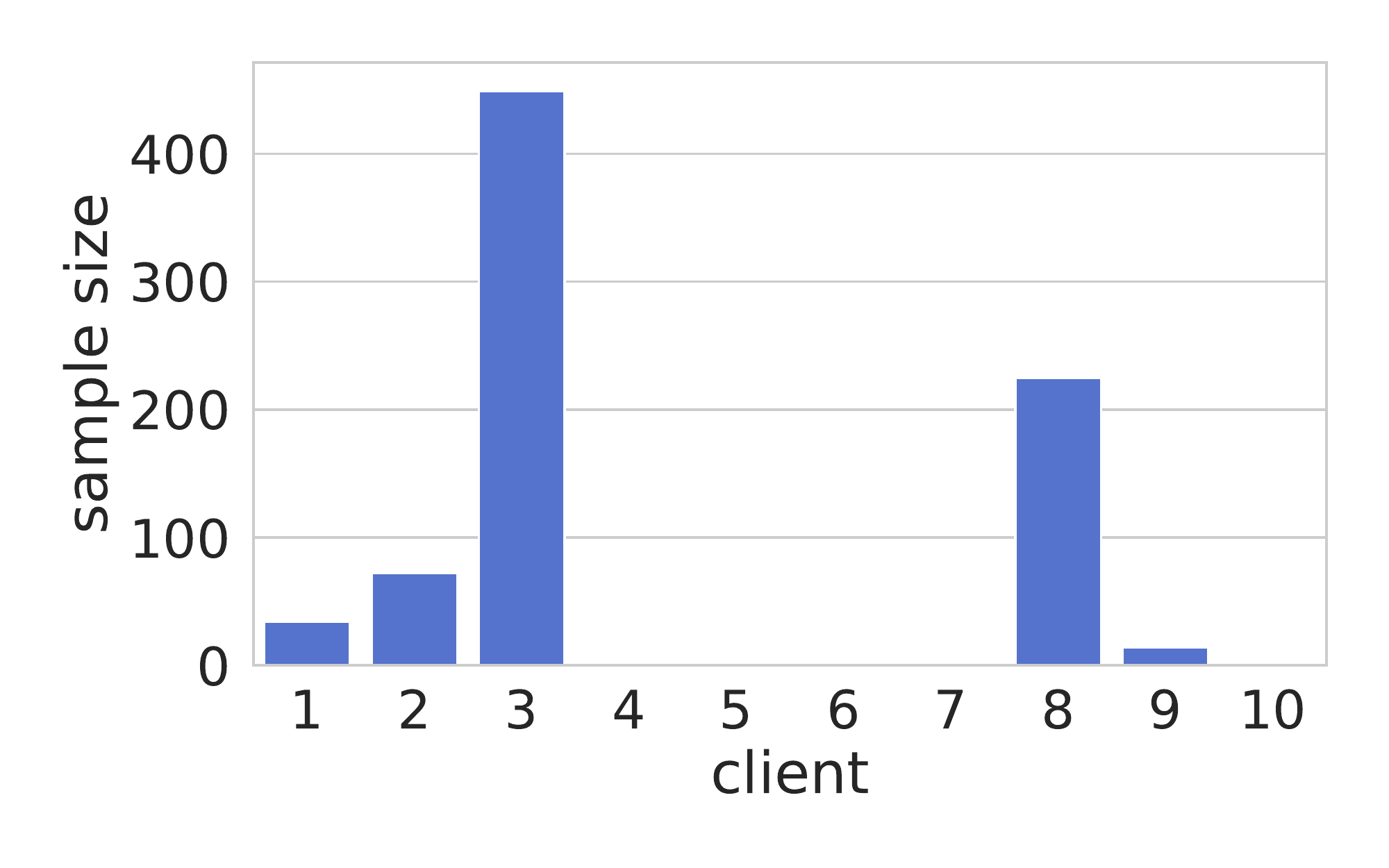}}}
\subfigure[\label{fig:alpha10} $\alpha = 10.0$]
{{\includegraphics[width=0.4\textwidth]{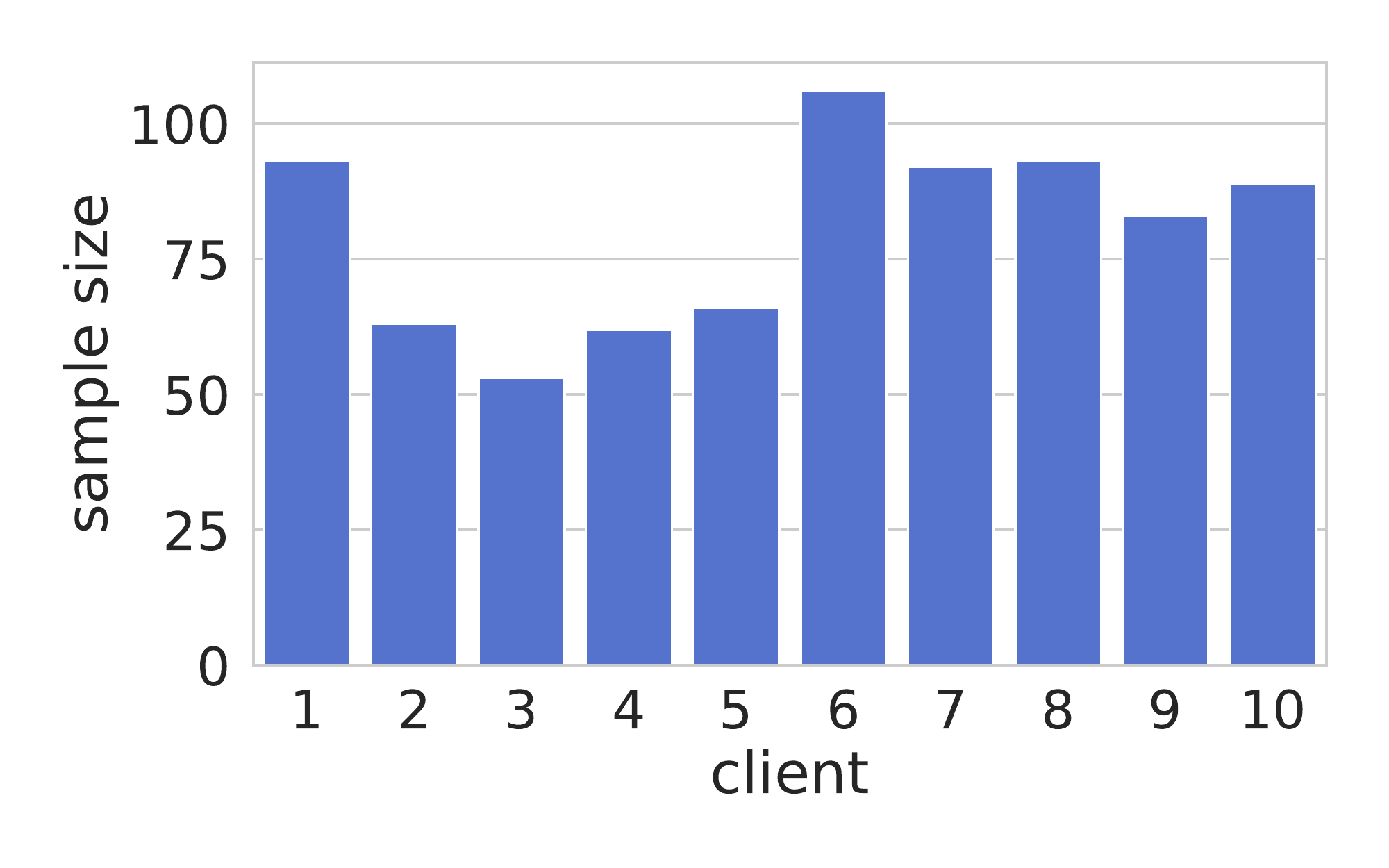}}}
\caption{Unbalanced sample distribution (non-I.I.D.) for citation networks.}
\label{fig:noniid_dists_citation}
\end{figure}

\begin{figure}[h!]
\centering
\subfigure[\label{fig:bace} BACE (\#clients: 4, alpha: 0.5)] {{\includegraphics[width=0.40\textwidth]{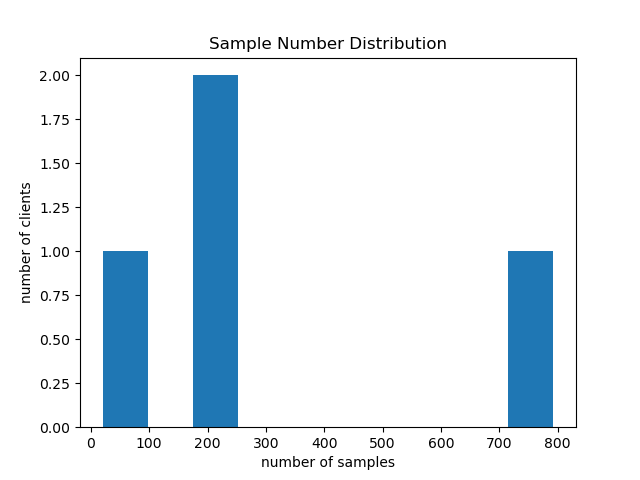}}}
\subfigure[\label{fig:clintox} Clintox (\#clients: 4, alpha: 0.5)]
{{\includegraphics[width=0.40\textwidth]{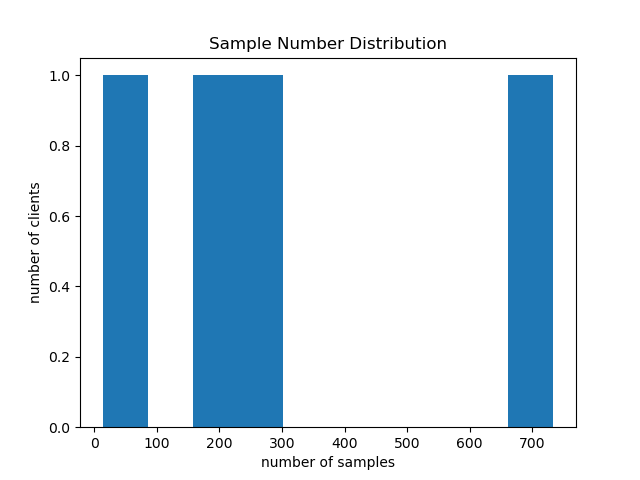}}}
\subfigure[\label{fig:PCBA} PCBA (\#clients: 8, alpha: 3)] {{\includegraphics[width=0.40\textwidth]{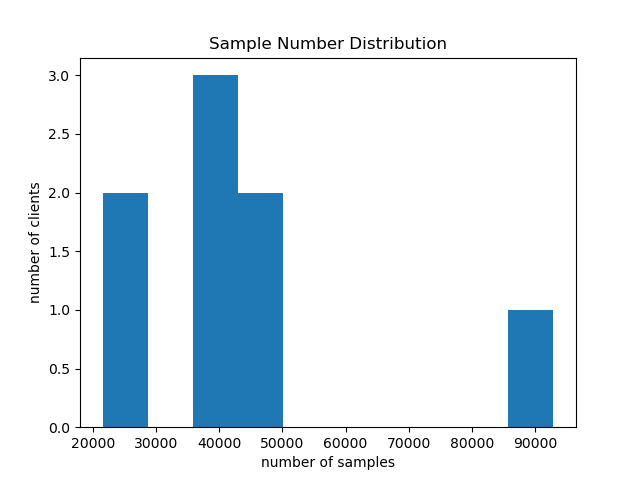}}}
\subfigure[\label{fig:tox21} Tox21 (\#clients: 8, alpha: 3)]
{{\includegraphics[width=0.40\textwidth]{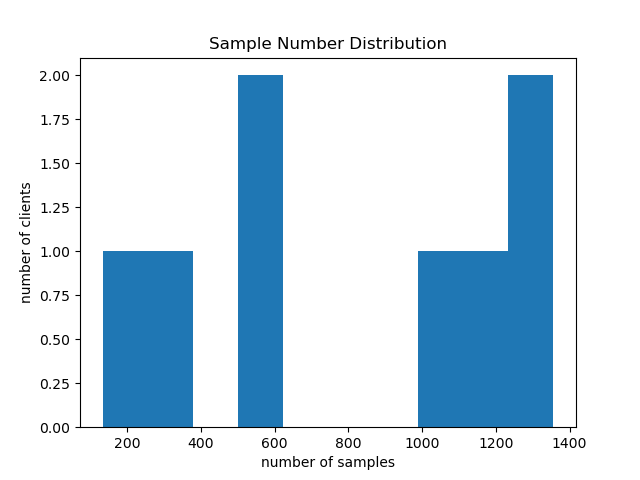}}}
\subfigure[\label{fig:bbbp} BBBP (\#clients: 4, alpha: 2)]
{{\includegraphics[width=0.40\textwidth]{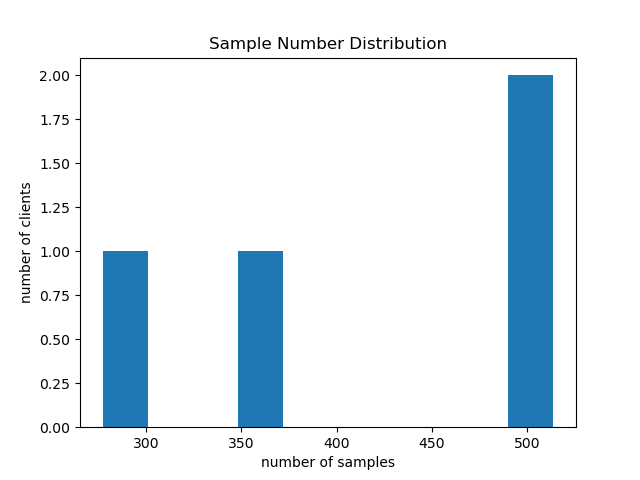}}}
\subfigure[\label{fig:SIDER} SIDER (\#clients: 4, alpha: 0.2)] {{\includegraphics[width=0.40\textwidth]{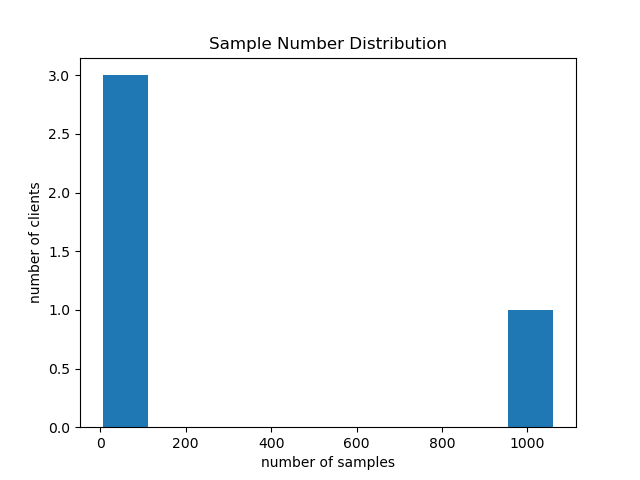}}}

\caption{\textcolor{black}{Unbalanced Sample Distribution (Non-I.I.D.) for Molecular Graph Classification Datasets}}
\label{fig:noniid_dists}
\end{figure}

\begin{figure}[h!]
\centering
\subfigure[\label{fig:herg-app} hERG (\#clients: 4, alpha: 3)] {{\includegraphics[width=0.40\textwidth]{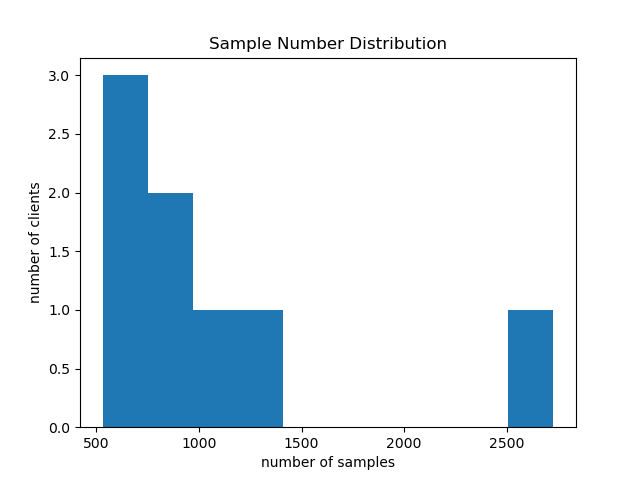}}}
\subfigure[\label{fig:esol} ESOL (\#clients: 4, alpha: 2)]
{{\includegraphics[width=0.40\textwidth]{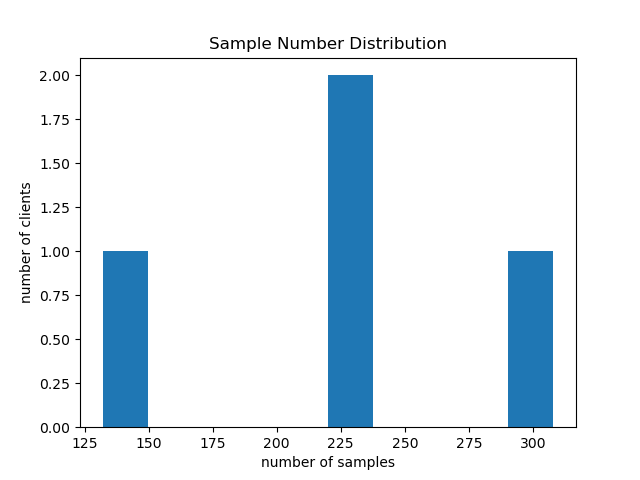}}}

\subfigure[\label{fig:qm9-app} QM9 (\#clients: 8, alpha: 3)]
{{\includegraphics[width=0.40\textwidth]{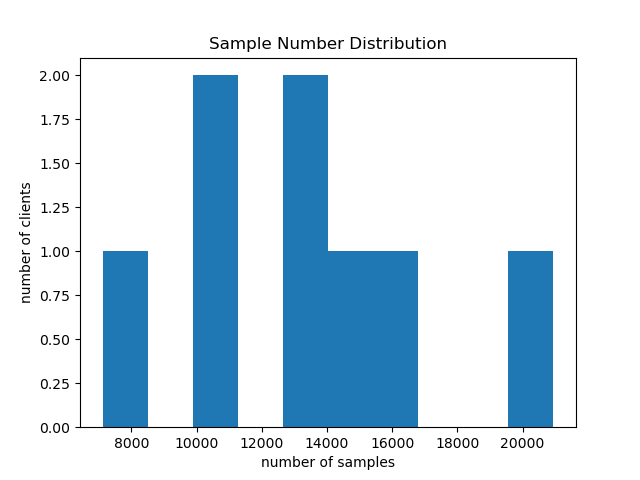}}}
\subfigure[\label{fig:lipo} LIPO (\#clients: 8, alpha: 2)]
{{\includegraphics[width=0.45\textwidth]{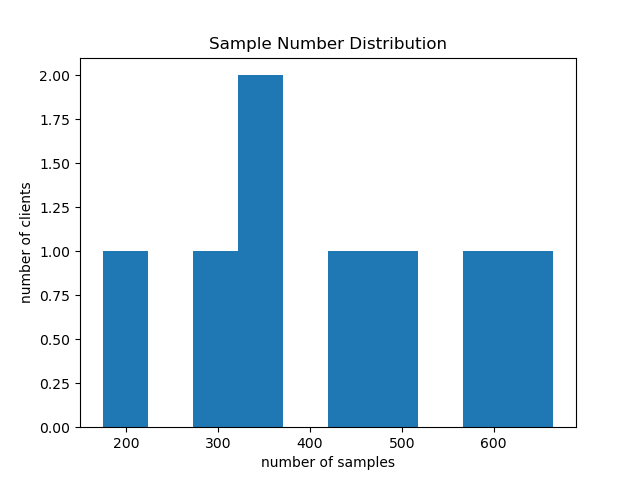}}}
\caption{\textcolor{black}{Example (Non-I.I.D.) Sample Distributions  for Molecular Property Prediction Datasets}}
\label{fig:noniid_dists_reg}
\end{figure}

\begin{figure}[h!]
\centering
\subfigure[\label{fig:ciao_distribution} Ciao ] {{\includegraphics[width=0.4\textwidth]{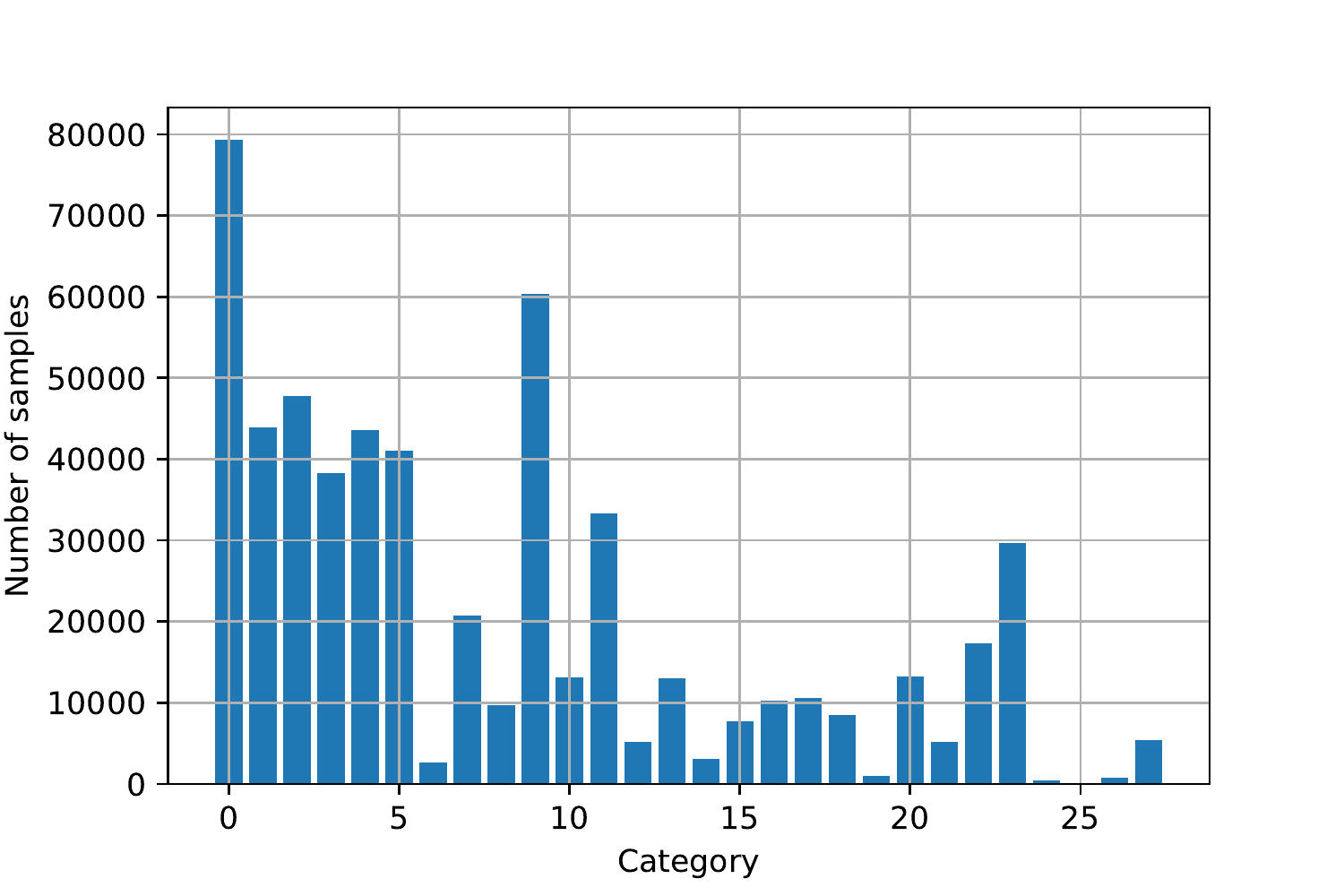}}}
\subfigure[\label{fig:epinions_distribution} Epinions]
{{\includegraphics[width=0.4\textwidth]{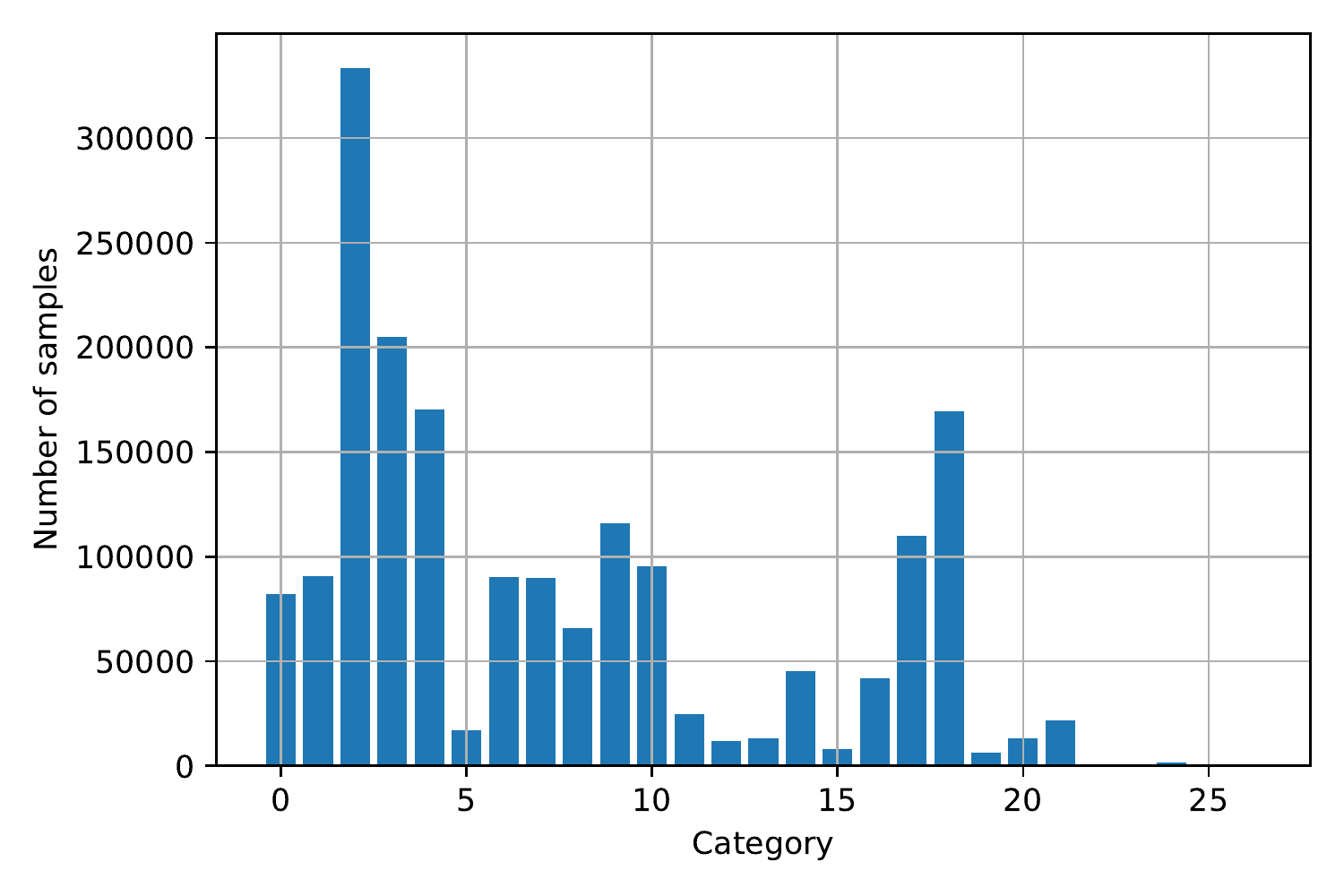}}}

\caption{\textcolor{black}{Example (Non-I.I.D.) Sample Distributions for Recommendation Systems Datasets}}
\label{fig:noniid_dists_recommendation}
\end{figure}

\paragraph{Based on the meta-data of dataset}
Data can be partitioned to clients by meta-data information. Meta-data can show the intrinsic data partition in real life scenes. For example, in recommendation system, user's behaviour is different for items from different categories. Fig. \ref{fig:noniid_dists_recommendation} shows the non-I.I.D. of user's rating number on item from different categories.

\clearpage
\section{More Details of \ours System Design}
\label{app:system}

 \texttt{Data Collector and Manager} is a distributed computing system that can collect scattered datasets or features from multiple servers to \texttt{Training Manager}. Such collection can also keep the raw data in the original server with RPCs, which can only access the data during training. After obtaining all necessary datasets for federated training, \texttt{Training Manager} will start federated training using algorithms supported by \texttt{FedML-API}. Once training has been completed, \texttt{Model Serving} can request the trained model to deploy for inference. Under this SDK abstraction, we plan to address the challenges mentioned above (1) and (2) within the \texttt{Data Collector and Manager}. As for challenge (3), we plan to make \texttt{FedML Client SDK} compatible with any operating systems (Linux, Android, iOS) with a cross-platform abstraction interface design.  In essence, the three modules inside \texttt{FedML Client SDK} builds up a pipeline that manages a model's life cycle, from federated training to personalized model serving (inference). Unifying three modules of a pipeline into a single SDK can simplify the system design. Any subsystem in an institute can integrate \texttt{FedML Client SDK} with a host process, which can be the backend service or desktop application. Overall, we hope \texttt{FedML Client SDK} could be a lightweight and easy-to-use SDK for federated learning among diverse cross-silo institutes.
 
\section{More Results of System Efficiency and Security}

\subsection{Evaluation on System Efficiency}
\label{app:system-analysis}

\begin{table}[htbp]
\centering
 \caption{System-Level Performance Metrics for Graph-Level \ours tasks with FedAvg (Hardware: 8 x NVIDIA Quadro RTX 5000 GPU (16GB/GPU); RAM: 512G; CPU: Intel Xeon Gold 5220R 2.20GHz).}
\label{tab:perf-results-graph}
\resizebox{ \linewidth}{!}{
\begin{threeparttable}
    \begin{tabular}{lccccccccccc}
    \toprule
      \textbf{} & & \textbf{SIDER} & \textbf{BACE} & \textbf{Clintox} & \textbf{BBBP} & \textbf{Tox21} & \textbf{FreeSolv} & \textbf{ESOL} & \textbf{Lipo} & \textbf{hERG} & \textbf{QM9}\\
      \midrule
        \multirow{3}{*}{Wall-clock Time} & GCN  & 5m 58s & 4m 57s & 4m 40s & 4m 13s & 15m 3s & 4m 12s & 5m 25s & 16m 14s & 35m 30s & 6h 48m \\
      & GAT  & 8m 48s & 5m 27s & 7m 37s & 5m 28s & 25m 49s & 6m 24s & 8m 36s & 25m 28s & 58m 14s & 9h 21m \\
      & GraphSAGE  & 2m 7s & 3m 58s& 4m 42s & 3m 26s & 14m 31s & 5m 53s & 6m 54s & 15m 28s & 32m 57s & 5h 33m \\
      \midrule
       \multirow{3}{*}{Average FLOP} & GCN & 697.3K & 605.1K & 466.2K & 427.2K & 345.8K & 142.6K & 231.6K & 480.6K & 516.6K & 153.9K \\
      & GAT & 703.4K  & 612.1K & 470.2K & 431K & 347.8K & 142.5K & 232.6K & 485K & 521.3K & 154.3K \\
      & GraphSAGE & 846K & 758.6K & 1.1M & 980K & 760.6K & 326.9K & 531.1K & 1.5M  & 1.184M & 338.2K \\
      \midrule
      \multirow{3}{*}{Parameters} & GCN  & 15.1K  & 13.5K & 13.6K & 13.5K & 14.2K & 13.5K & 13.5K &13.5K & 13.5K & 14.2K\\
      & GAT & 20.2K & 18.5K & 18.6K & 18.5K & 19.2K & 18.5K & 18.5K & 18.5K & 18.5K & 19.2K\\
      & GraphSAGE & 10.6K & 8.9K & 18.2K & 18.1K & 18.8K & 18.1K & 18.1K & 269K & 18.1K & 18.8K \\
       \bottomrule
    \end{tabular}
    \begin{tablenotes}[para,flushleft]
      \footnotesize
      \item *Note that we use the distributed training paradigm where each client's local training uses one GPU. Please refer to our code for details.
    \end{tablenotes}
\end{threeparttable}}
\label{table:system-analysis-graph}
\end{table}

\begin{table}[htbp]
\centering
 \caption{System-level Performance Metrics for Subgraph-Level \ours tasks with FedAvg (Hardware: 8 x NVIDIA Quadro RTX 5000 GPU (16GB/GPU); RAM: 512G; CPU: Intel Xeon Gold 5220R 2.20GHz).}
\label{tab:perf-results-subgraph}
\resizebox{ 0.5\columnwidth}{!}{
\begin{threeparttable}
    \begin{tabular}{lccc}
    \toprule
      \textbf{} & & \textbf{Ciao} & \textbf{Epinions} \\
      \midrule
        \multirow{3}{*}{Wall-clock Time} & GCN  & 352s & 717s \\
      & GAT  & 350s & 749s\\
      & GraphSAGE  & 551s & 810s \\
      \midrule
       \multirow{3}{*}{Average FLOP} & GCN & 697.3K & 605.1K \\
      & GAT & 703.4K  & 612.1K \\
      & GraphSAGE & 846K & 758.6K  \\
      \midrule
      \multirow{3}{*}{Parameters} & GCN  & 15.1K  & 13.5K \\
      & GAT & 20.2K & 18.5K \\
      & GraphSAGE & 10.6K & 8.9K  \\
       \bottomrule
    \end{tabular}
    \begin{tablenotes}[para,flushleft]
      \footnotesize
      \item *Note that we use the distributed training paradigm where each client's local training uses one GPU. Please refer to our code for details.
    \end{tablenotes}
\end{threeparttable}}
\label{table:system-analysis-subgraph}
\end{table}
 
 \begin{table}[htbp]
\centering
 \caption{System-level Performance Metrics for Node-Level \ours tasks with FedAvg (Hardware: 8 x NVIDIA Quadro RTX 5000 GPU (16GB/GPU); RAM: 512G; CPU: Intel Xeon Gold 5220R 2.20GHz).}
\label{tab:perf-results-node}
\resizebox{ 0.65\columnwidth}{!}{
\begin{threeparttable}
    \begin{tabular}{lccccc}
    \toprule
      \textbf{} & & \textbf{CORA} & \textbf{Citeseer} & \textbf{DBLP} & \textbf{PubMed} \\
      \midrule
        \multirow{3}{*}{Wall-clock Time} & GCN  & 833s & 622s & 654s & 653s \\
      & GAT  & 871s & 652s & 682s & 712s \\
      & GraphSAGE  & 774s & 562s & 622s &  592s\\
      \midrule
       \multirow{3}{*}{Average FLOP} & GCN & 44.9M & 739.4K   & 8.8M & 1.9M \\
      & GAT & 47.8M  & 845.3K & 9.5M & 2.3M\\
      & GraphSAGE & 45.3M & 817K  & 9.2M & 2.1M\\
      \midrule
      \multirow{3}{*}{Parameters} & GCN  & 282.1K  & 20.5K &  53.7K&  17.2K\\
      & GAT & 285.1K & 23.7K & 56.5K & 19.3K \\
      & GraphSAGE & 283K & 21.6K & 54.7K & 18.2K \\
       \bottomrule
    \end{tabular}
    \begin{tablenotes}[para,flushleft]
      \footnotesize
      \item *Note that we use the distributed training paradigm where each client's local training uses one GPU. Please refer to our code for details.
    \end{tablenotes}
\end{threeparttable}}
\label{table:system-analysis-node}
\end{table}

The training time using RPC is also evaluated; and results are similar to that of using MPI. Note that RPC is useful for realistic deployment when GPU/CPU-based edge devices can only be accessed via public IP addresses due to locating in different data centers. We will provide detailed test results in such a scenario in our future work.

\subsection{Evaluation on Security (\texttt{LightSecAgg})}
\label{app:lightsecagg}

\begin{figure}[h!]
    \centering
    \subfigure[Non-overlapped]{\label{fig:runtime_CNN_NonOverlap_varDropout}
    \includegraphics[width=.47\textwidth]{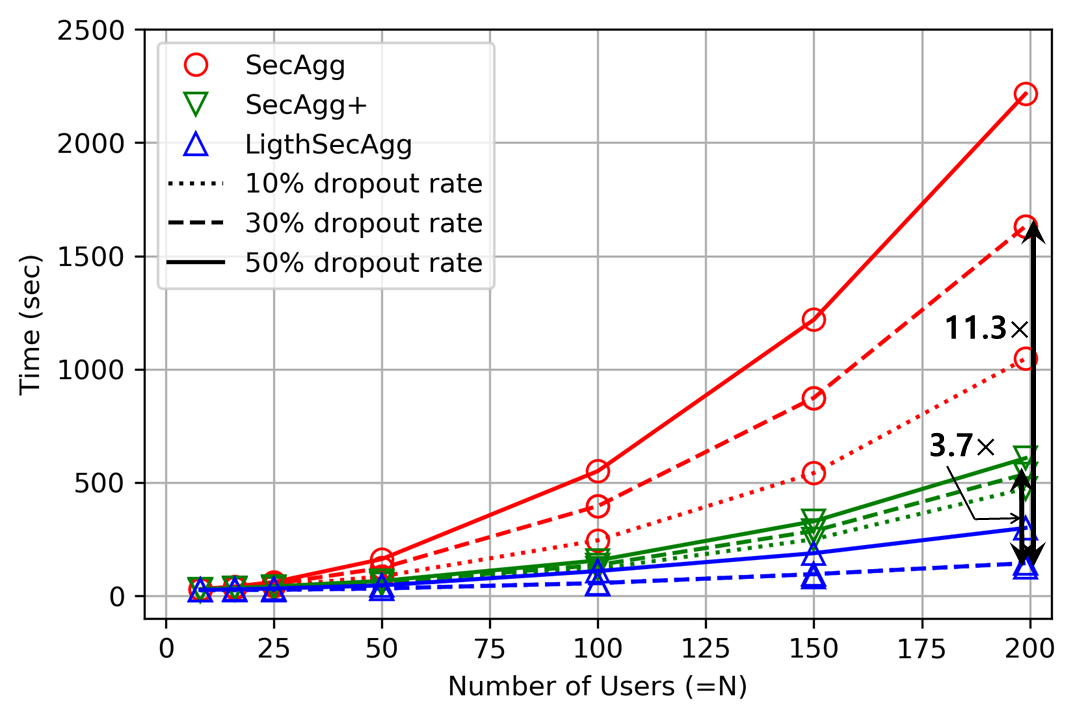}
    }
    \subfigure[Overlapped]{\label{fig:runtime_CNN_Overlap_varDropout}
    \includegraphics[width=.47\textwidth]{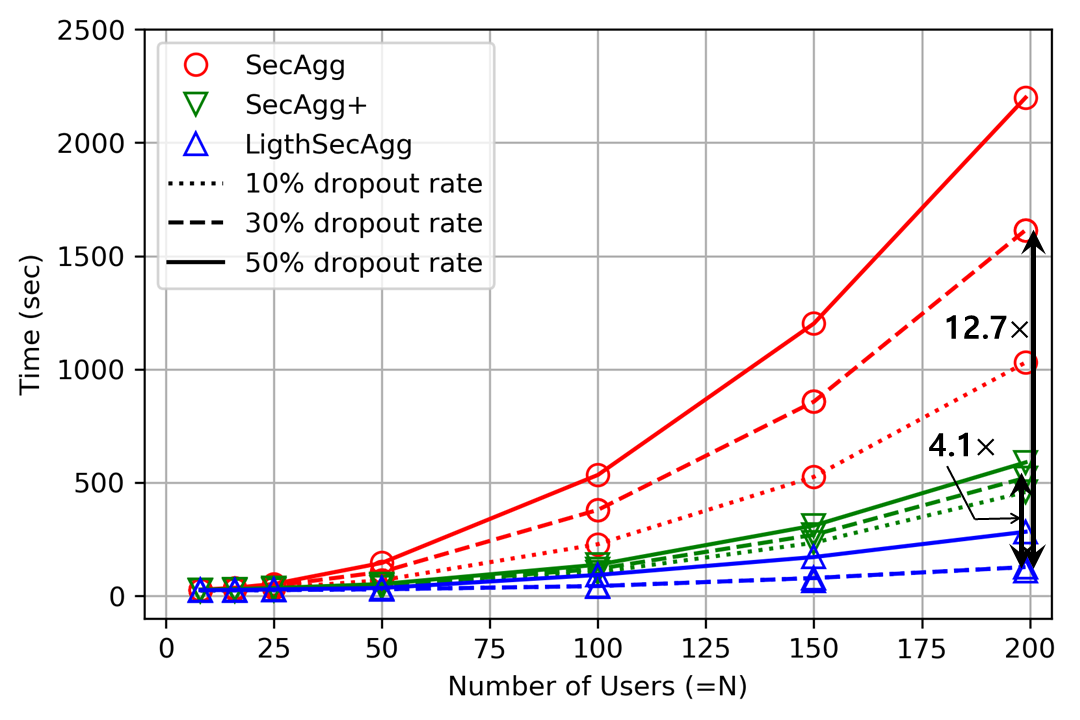}
    }
\caption{\texttt{LightSecAgg}: Total running time of \texttt{LightSecAgg} versus the state-of-the-art protocols (SecAgg \cite{bonawitz2017practical} and SecAgg+ \cite{bell2020secure}) to train neural networks with 1.2 M parameters (\textit{more than all GNN models used in this work}), as the number of users increases, for various dropout rates.}
\label{fig:runtime_CNN_varDropout}
\end{figure}

\texttt{LightSecAgg}, a FL security algorithm developed by our team \footnote{\texttt{LightSecAgg} is under submission and patent application process when we publish \ours, so we cannot provide a full paper reference now. At the current stage, to understand secure aggregation, we refer readers to the baseline SecAgg \cite{bonawitz2017practical}.}, provides model privacy guarantees as the state-of-the-art (SecAgg \cite{bonawitz2017practical} and SecAgg+ \cite{bell2020secure}) while substantially reducing the aggregation (hence run-time) complexity (Figure \ref{fig:runtime_CNN_varDropout}). The main idea of \texttt{LightSecAgg} is that each user protects its local model using a locally generated random mask. This mask is then encoded and shared to other users, in such a way that the aggregate mask of any sufficiently large set of surviving users can be directly reconstructed at the server. Our main effort in \ours is integrating  \texttt{LightSecAgg}, optimizing its system performance, and designing user-friendly APIs to make it compatible with various models and FL algorithms.

\clearpage
 
\section{More Details of the Empirical Analysis}
\label{app:experiments}

\subsection{Hyper-parameters}
\label{app:experiments:hp}
For each task, we utilize grid search to find the best results. Table~\ref{tab:hyper-parameters} \& \ref{tab:hyper-parameters-fed} list all the hyper-parameters ranges used in our experiments. All hyper-parameter tuning is run on a single GPU. The best hyperparameters for each dataset and model are listed in Table \ref{tab:BBBP_Tox21_ToxCast_SIDER},\ref{tab:BBBP_Tox21_ToxCast_SIDER_fed},\ref{tab:ESOL_FreeSolv_Lipo}, \& \ref{tab:ESOL_FreeSolv_Lipo_fed} For molecule tasks ,batch-size is kept fixed since the molecule-level task requires us to have mini-batch is equal to 1. Also, number of GNN layers were fixed to 2 because having too many GNN layers result in over-smoothing phenomenon as shown in \citep{li2018deeper}. For all experiments, we used Adam optimizer.

\begin{table}[htbp]
  \centering
  \caption{Hyper-parameter Range for Graph-Level Centralized Training(classification \& regression) }
  \resizebox{1\textwidth}{!}{
    \begin{tabular}{l|l|l}
    \toprule
    hyper-parameter & Description & Range \\
    \midrule
    learning rate & Rate of speed at which the model learns.  & $\left[0.00015, 0.0015, 0.015, 0.15 \right]$ \\
    dropout rate & Dropout ratio & $\left[0.2, 0.3, 0.5 ,0.6 \right]$ \\
    node embedding dimension & Dimensionality of the node embedding & $\left[16,32,64, 128, 256 \right]$ \\
    hidden layer dimension & Hidden layer dimensionality &  $\left[16,32,64,128, 256 \right]$ \\
    readout embedding dimension & Dimensionality of the readout embedding &  $\left[16,32,64,128 256 \right]$ \\
    graph embedding dimension & Dimensionality of the graph embedding &  $\left[16,32,64,128, 256 \right]$ \\
    attention heads & Number of attention heads required for GAT & 1-7 \\
    alpha & LeakyRELU parameter used in GAT model & 0.2  \\
    \bottomrule
    \end{tabular}
    }
  \label{tab:hyper-parameters}%
\end{table}

\begin{table}[htbp]
  \centering
  \caption{Hyper-parameter Range for Graph-Level Federated Learning(classification \& regression)}
  \resizebox{1\textwidth}{!}{
    \begin{tabular}{l|l|l}
    \toprule
    hyper-parameter & Description & Range \\
    \midrule
    learning rate & Rate of speed at which the model learns.  & $\left[0.00015, 0.0015, 0.015, 0.15 \right]$ \\
    dropout rate & Dropout ratio & $\left[ 0.3, 0.5, 0.6 \right]$ \\
    node embedding dimension & Dimensionality of the node embedding & 64 \\
    hidden layer dimension & Hidden layer dimensionality &  64 \\
    readout embedding dimension & Dimensionality of the readout embedding &  64 \\
    graph embedding dimension & Dimensionality of the graph embedding &  64 \\
    attention heads & Number of attention heads required for GAT & 1-7 \\
    alpha & LeakyRELU parameter used in GAT model & 0.2  \\
    rounds & Number of federating learning rounds & [10, 50, 100] \\
    epoch  & Epoch of  clients & 1 \\
    number of clients & Number of users in a federated learning round & 4-10 \\
    \bottomrule
    \end{tabular}
    }
  \label{tab:hyper-parameters-graph-fed}
\end{table}%

\begin{table}[htbp]
  \centering
  \caption{Hyper-parameter Range for Subgraph-Level Federated Learning}
  \resizebox{1\textwidth}{!}{
    \begin{tabular}{l|l|l}
    \toprule
    hyper-parameter & Description & Range \\
    \midrule
    learning rate & Rate of speed at which the model learns.  & $\left[0.0001, 0.001, 0.01, 0.1 \right]$ \\
    node embedding dimension & Dimensionality of the node embedding & 64 \\
    hidden layer dimension & Hidden layer dimensionality &  $\left[64 \right]$ \\
    rounds & Number of federating learning rounds & $\left[1, 10, 20, 50, 100 \right]$ \\
    local epoch  & Epoch of  clients & $\left[1, 2, 5 \right]$ \\
    number of clients & Number of users in a federated learning round & 4-10 \\
    \bottomrule
    \end{tabular}
    }
  \label{tab:hyper-parameters-subgraph-fed}
\end{table}%

\begin{table}[htbp]
  \centering
  \caption{Hyper-parameter Range for Node-Level Federated Learning}
  \resizebox{1\textwidth}{!}{
    \begin{tabular}{l|l|l}
    \toprule
    hyper-parameter & Description & Range \\
    \midrule
    learning rate & Rate of speed at which the model learns.  & $\left[0.1, 0.01, 0.001, 0.0001 \right]$ \\
    dropout rate & Dropout ratio & $\left[ 0.3, 0.5, 0.6 \right]$ \\
    hidden layer dimension & Hidden layer dimensionality &  [32, 64, 128] \\
    alpha & LeakyRELU parameter used in GAT model & [0.1, 10]  \\
    rounds & Number of federating learning rounds & 100 \\
    epoch  & Epoch of  clients & [1, 3, 5] \\
    number of clients & Number of users in a federated learning round & 10 \\
    \bottomrule
    \end{tabular}
    }
  \label{tab:hyper-parameters-node-fed}
\end{table}%

\begin{table}[h!]
\label{tab:clf-params}
    \small
    \centering
    \caption{Hyperparameters for Graph-Level Molecular Classification Task}
    \begin{tabular}{ |c|c|c|c|c| } 
    \hline
    \textbf{Dataset} & \textbf{Score \& Parameters} & \textbf{GCN} & \textbf{GAT} & \textbf{GraphSAGE} \\
    \hline
    \hline    
    \multirow{9}{*}{BBBP}
    & ROC-AUC Score & 0.8705 & 0.8824  & \textbf{0.8930} \\\cline{2-5}
    & learning rate & 0.0015 & 0.015  & 0.01 \\\cline{2-5}
    & dropout rate & 0.2 & 0.5 & 0.2 \\\cline{2-5}
    & node embedding dimension & 64 & 64 &  64 \\\cline{2-5}
    & hidden layer dimension & 64 & 64 & 64 \\\cline{2-5}
    & readout embedding dimension & 64 & 64 & 64 \\\cline{2-5}
    & graph embedding dimension & 64 & 64 & 64 \\\cline{2-5}
    & attention heads & None & 2 &  None \\\cline{2-5}
    & alpha & None & 0.2 &  None \\\cline{2-5}
    \hline
    \multirow{9}{*}{BACE}
    & ROC-AUC Score & 0.9221 & 0.7657 & \textbf{0.9266} \\\cline{2-5}
    & learning rate & 0.0015 & 0.001 & 0.0015 \\\cline{2-5}
    & dropout rate &0.3 & 0.3 & 0.3 \\\cline{2-5}
    & node embedding dimension & 64 & 64 & 16 \\\cline{2-5}
    & hidden layer dimension & 64 & 64 & 64 \\\cline{2-5}
    & readout embedding dimension & 64 & 64 & 64 \\\cline{2-5}
    & graph embedding  dimension & 64 & 64 & 64 \\\cline{2-5}
    & attention heads & None & 2 &  None \\\cline{2-5}
    & alpha & None & 0.2  & None \\\cline{2-5}
    \hline   
    \multirow{9}{*}{Tox21}
    & ROC-AUC Score &  0.7800 & 0.8144 &  \textbf{0.8317} \\\cline{2-5}
    & learning rate & 0.0015 & 0.00015 &  0.00015 \\\cline{2-5}
    & dropout rate  & 0.4 & 0.3 & 0.3 \\\cline{2-5}
    & node embedding dimension & 64 & 128 &  256 \\\cline{2-5}
    & hidden layer dimension & 64 & 64 &  128 \\\cline{2-5}
    & readout embedding dimension & 64 & 128 &  256 \\\cline{2-5}
    & graph embedding dimension&  64 & 64 &  128 \\\cline{2-5}
    & attention heads &  None & 2 &  None \\\cline{2-5}
    & alpha & None & 0.2 &  None \\\cline{2-5}
    \hline    
    \multirow{9}{*}{SIDER}
    & ROC-AUC Score & 0.6476 & 0.6639 & \textbf{0.6669} \\\cline{2-5}
    & learning rate & 0.0015 & 0.0015 &  0.0015 \\\cline{2-5}
    & dropout rate & 0.3 & 0.3 &  0.6 \\\cline{2-5}
    & node embedding dimension & 64 & 64 &  16 \\\cline{2-5}
    & hidden layer dimension & 64 & 64 &  64 \\\cline{2-5}
    & readout embedding dimension &  64 & 64 &  64 \\\cline{2-5}
    & graph embedding dimension & 64 & 64 & 64 \\\cline{2-5}
    & attention heads & None & 2 & None \\\cline{2-5}
    & alpha & None & 0.2 &  None \\\cline{2-5}
    \hline       
    \multirow{9}{*}{ClinTox}
     & ROC-AUC Score & 0.8914 & 0.9573 & \textbf{0.9716} \\\cline{2-5}
    & learning rate & 0.0015 & 0.0015 &  0.0015 \\\cline{2-5}
    & dropout rate & 0.3 & 0.3 &  0.3 \\\cline{2-5}
    & node embedding dimension & 64 & 64 &  64 \\\cline{2-5}
    & hidden layer dimension & 64 & 64 &  64 \\\cline{2-5}
    & readout embedding dimension & 64 & 64 &  64 \\\cline{2-5}
    & graph embedding dimension & 64 & 64 & 64 \\\cline{2-5}
    & attention heads &  None & 2 & None \\\cline{2-5}
    & alpha & None & 0.2 &  None \\\cline{2-5}
    \hline    
    \end{tabular}
    \label{tab:BBBP_Tox21_ToxCast_SIDER}
\end{table}

\begin{table}[h!]
\label{tab:fedclf-params}
    \small
    \centering
    \caption{Hyperparameters for Graph-Level Federated Molecular Classification Task}
    \begin{tabular}{ |c|c|c|c|c| } 
    \hline
    \textbf{Dataset} & \textbf{Score \& Parameters} & \textbf{GCN + FedAvg} & \textbf{GAT + FedAvg} & \textbf{GraphSAGE + FedAvg} \\
    \hline
    \hline   
     \multirow{10}{*}{BBBP}
    & ROC-AUC Score & 0.7629 & 0.8746  & \textbf{0.8935} \\\cline{2-5}
    & number of clients & 4 & 4 & 4 \\\cline{2-5}
    & learning rate & 0.0015 & 0.0015 & 0.015 \\\cline{2-5}
    & dropout rate & 0.3 & 0.3 &  0.6 \\\cline{2-5}
    & node embedding dimension & 64 & 64 & 64 \\\cline{2-5}
    & hidden layer dimension & 64 & 64 & 64 \\\cline{2-5}
    & readout embedding dimension & 64 & 64 & 64 \\\cline{2-5}
    & graph embedding dimension & 64 & 64 & 64 \\\cline{2-5}
    & attention heads & None & 2 &  None \\\cline{2-5}
    & alpha & None & 0.2 &  None \\\cline{2-5}
    \hline    
    \multirow{10}{*}{BACE}
    & ROC-AUC Score & 0.6594 & 0.7714  & \textbf{0.8604} \\\cline{2-5}
    & number of clients & 4 & 4 & 4 \\\cline{2-5}
    & learning rate & 0.0015 & 0.0015 & 0.0015 \\\cline{2-5}
    & dropout rate & 0.5 & 0.3 & 0.5 \\\cline{2-5}
    & node embedding dimension & 64 & 64 & 16 \\\cline{2-5}
    & hidden layer dimension & 64 & 64 & 64 \\\cline{2-5}
    & readout embedding dimension & 64 & 64 & 64 \\\cline{2-5}
    & graph embedding dimension & 64 & 64 & 64 \\\cline{2-5}
    & attention heads & None & 2 &  None \\\cline{2-5}
    & alpha & None & 0.2 &  None \\\cline{2-5}
    \hline    
   
    \multirow{10}{*}{Tox21}
    & ROC-AUC Score & 0.7128 & 0.7171  & \textbf{0.7801} \\\cline{2-5}
    & number of clients & 4 & 4 & 4 \\\cline{2-5}
    & learning rate & 0.0015 & 0.0015 & 0.00015 \\\cline{2-5}
    & dropout rate & 0.6 & 0.3 & 0.3 \\\cline{2-5}
    & node embedding dimension & 64 & 64 & 64 \\\cline{2-5}
    & hidden layer dimension & 64 & 64& 64 \\\cline{2-5}
    & readout embedding dimension & 64 & 64 & 64 \\\cline{2-5}
    & graph embedding dimension & 64 & 64 & 64 \\\cline{2-5}
    & attention heads & None & 2 &  None \\\cline{2-5}
    & alpha & None & 0.2 &  None \\\cline{2-5}
    \hline
    \multirow{10}{*}{SIDER}
    & ROC-AUC Score & 0.6266 & 0.6591  & \textbf{0.67} \\\cline{2-5}
    & number of clients & 4 & 4 & 4 \\\cline{2-5}
    & learning rate & 0.0015 & 0.0015 & 0.0015 \\\cline{2-5}
    & dropout rate & 0.6 & 0.3 & 0.6 \\\cline{2-5}
    & node embedding dimension & 64 & 64 & 16 \\\cline{2-5}
    & hidden layer dimension & 64 & 64 & 64 \\\cline{2-5}
    & readout embedding dimension & 64 & 64 & 64 \\\cline{2-5}
    & graph embedding dimension & 64 & 64 & 64 \\\cline{2-5}
    & attention heads & None & 2 &  None \\\cline{2-5}
    & alpha & None & 0.2 &  None \\\cline{2-5}
    \hline
    \multirow{7}{*}{ClinTox}
    & ROC-AUC Score & 0.8784 & 0.9160  & \textbf{0.9246} \\\cline{2-5}
    & number of clients & 4 & 4 & 4 \\\cline{2-5}
    & learning rate & 0.0015 & 0.0015 & 0.015 \\\cline{2-5}
    & dropout rate & 0.5 & 0.6 & 0.3 \\\cline{2-5}
    & node embedding dimension & 64 & 64 & 64 \\\cline{2-5}
    & hidden layer dimension & 64 & 64 & 64 \\\cline{2-5}
    & readout embedding dimension & 64 & 64 & 64 \\\cline{2-5}
    & graph embedding dimension & 64 & 64 & 64 \\\cline{2-5}
    & attention heads & None & 2 &  None \\\cline{2-5}
    & alpha & None & 0.2 &  None \\\cline{2-5}
    \hline
    \end{tabular}
    \label{tab:BBBP_Tox21_ToxCast_SIDER_fed}
\end{table}
    
\begin{table}[h!]
\label{tab:reg-params}
    \small
    \centering
    \caption{Hyperparameters for Graph-Level Molecular Regression Task}
    \begin{tabular}{ |c|c|c|c|c| } 
    \hline
    \textbf{Dataset} & \textbf{Score \&Parameters} & \textbf{GCN} & \textbf{GAT} & \textbf{GraphSAGE} \\
    \hline
    \hline    
    \multirow{9}{*}{Freesolv}
     & RMSE Score & 0.8705 & 0.8824  & \textbf{0.8930} \\\cline{2-5}
    & learning rate & 0.0015 & 0.015  & 0.01 \\\cline{2-5}
    & dropout rate & 0.2 & 0.5 & 0.2 \\\cline{2-5}
    & node embedding dimension & 64 & 64 &  64 \\\cline{2-5}
    & hidden layer dimension & 64 & 64 & 64 \\\cline{2-5}
    & readout embedding dimension & 64 & 64 & 64 \\\cline{2-5}
    & graph embedding dimension & 64 & 64 & 64 \\\cline{2-5}
    & attention heads & None & 2 &  None \\\cline{2-5}
    & alpha & None & 0.2 &  None \\\cline{2-5}
    \hline
 
    \multirow{9}{*}{ESOL}
    & RMSE Score & 0.8705 & 0.8824  & \textbf{0.8930} \\\cline{2-5}
    & learning rate & 0.0015 & 0.015  & 0.01 \\\cline{2-5}
    & dropout rate & 0.2 & 0.5 & 0.2 \\\cline{2-5}
    & node embedding dimension & 64 & 64 &  64 \\\cline{2-5}
    & hidden layer dimension & 64 & 64 & 64 \\\cline{2-5}
    & readout embedding dimension & 64 & 64 & 64 \\\cline{2-5}
    & graph embedding dimension & 64 & 64 & 64 \\\cline{2-5}
    & attention heads & None & 2 &  None \\\cline{2-5}
    & alpha & None & 0.2 &  None \\\cline{2-5}
    \hline
  
    \multirow{9}{*}{Lipophilicity}
    & RMSE Score & 0.8521 & 0.7415 & \textbf{0.7078} \\\cline{2-5}
    & learning rate & 0.0015 & 0.001  & 0.001 \\\cline{2-5}
    & dropout rate & 0.3 & 0.3 & 0.3 \\\cline{2-5}
    & node embedding dimension & 128 & 128 &  128 \\\cline{2-5}
    & hidden layer dimension & 64 & 64 & 64 \\\cline{2-5}
    & readout embedding dimension & 128 & 128 & 128 \\\cline{2-5}
    & graph embedding dimension & 64 & 64 & 64 \\\cline{2-5}
    & attention heads & None & 2 &  None \\\cline{2-5}
    & alpha & None & 0.2 &  None \\\cline{2-5}
    \hline
    
     \multirow{9}{*}{hERG}

    & RMSE Score & 0.7257 & \textbf{0.6271}  & 0.7132 \\\cline{2-5}
    & learning rate & 0.001 & 0.001  & 0.005 \\\cline{2-5}
    & dropout rate & 0.3 & 0.5 & 0.3 \\\cline{2-5}
    & node embedding dimension & 64 & 64 &  64 \\\cline{2-5}
    & hidden layer dimension & 64 & 64 & 64 \\\cline{2-5}
    & readout embedding dimension & 64 & 64 & 64 \\\cline{2-5}
    & graph embedding dimension & 64 & 64 & 64 \\\cline{2-5}
    & attention heads & None & 2 &  None \\\cline{2-5}
    & alpha & None & 0.2 &  None \\\cline{2-5}
    \hline
    
     \multirow{9}{*}{QM9}
    & RMSE Score & 14.78 & \textbf{12.44}  & 13.06 \\\cline{2-5}
    & learning rate & 0.0015 & 0.015  & 0.01 \\\cline{2-5}
    & dropout rate & 0.2 & 0.5 & 0.2 \\\cline{2-5}
    & node embedding dimension & 64 & 64 &  64 \\\cline{2-5}
    & hidden layer dimension & 64 & 64 & 64 \\\cline{2-5}
    & readout embedding dimension & 64 & 64 & 64 \\\cline{2-5}
    & graph embedding dimension & 64 & 64 & 64 \\\cline{2-5}
    & attention heads & None & 2 &  None \\\cline{2-5}
    & alpha & None & 0.2 &  None \\\cline{2-5}
    \hline
    \end{tabular}
    \label{tab:ESOL_FreeSolv_Lipo}
\end{table}

\begin{table}[h!]
\label{tab:fedreg-params}
    \small
    \centering
    \caption{Hyperparameters for Graph-Level Federated Molecular Regression Task}
    \begin{tabular}{ |c|c|c|c|c| } 
    \hline
    \textbf{Dataset} & \textbf{Parameters} & \textbf{GCN + FedAvg} & \textbf{GAT + FedAvg} & \textbf{GraphSAGE + FedAvg} \\
    \hline
    \hline    
    \multirow{10}{*}{FreeSolv}
     & RMSE Score & 2.747 & 3.108  & \textbf{1.641} \\\cline{2-5}
     & number of clients & 4 & 8 & 4 \\\cline{2-5}
    & learning rate & 0.0015 & 0.00015  & 0.015 \\\cline{2-5}
    & dropout rate & 0.6 & 0.5 & 0.6 \\\cline{2-5}
    & node embedding dimension & 64 & 64 &  64 \\\cline{2-5}
    & hidden layer dimension & 64 & 64 & 64 \\\cline{2-5}
    & readout embedding dimension & 64 & 64 & 64 \\\cline{2-5}
    & graph embedding dimension & 64 & 64 & 64 \\\cline{2-5}
    & attention heads & None & 2 &  None \\\cline{2-5}
    & alpha & None & 0.2 &  None \\\cline{2-5}
    \hline    
    \multirow{10}{*}{ESOL}
     & RMSE Score & 1.435 & \textbf{1.028}  & 1.185 \\\cline{2-5}
     & number of clients & 4 & 4 & 4 \\\cline{2-5}
    & learning rate & 0.0015 & 0.0015  & 0.0015 \\\cline{2-5}
    & dropout rate & 0.5 & 0.3 & 0.3 \\\cline{2-5}
    & node embedding dimension & 64 & 256 &  64 \\\cline{2-5}
    & hidden layer dimension & 64 & 64 & 64 \\\cline{2-5}
    & readout embedding dimension & 64 & 64 & 64 \\\cline{2-5}
    & graph embedding dimension & 64 & 64 & 64 \\\cline{2-5}
    & attention heads & None & 2 &  None \\\cline{2-5}
    & alpha & None & 0.2 &  None \\\cline{2-5}
    \hline    
    \multirow{10}{*}{Lipophilicity}
  & RMSE Score & 1.146 & 1.004  & \textbf{0.7788} \\\cline{2-5}
     & number of clients & 4 & 4 & 4 \\\cline{2-5}
    & learning rate & 0.0015 & 0.0015  & 0.0015 \\\cline{2-5}
    & dropout rate & 0.3 & 0.3 & 0.3 \\\cline{2-5}
    & node embedding dimension & 64 & 64 &  256 \\\cline{2-5}
    & hidden layer dimension & 64 & 64 & 256 \\\cline{2-5}
    & readout embedding dimension & 64 & 64 & 256 \\\cline{2-5}
    & graph embedding dimension & 64 & 64 & 256 \\\cline{2-5}
    & attention heads & None & 2 &  None \\\cline{2-5}
    & alpha & None & 0.2 &  None \\\cline{2-5}
    \hline
    \multirow{10}{*}{hERG}
  & RMSE Score & 0.7944 & 0.7322  & \textbf{0.7265} \\\cline{2-5}
     & number of clients & 8 & 8 & 8 \\\cline{2-5}
    & learning rate & 0.0015 & 0.0015  & 0.0015 \\\cline{2-5}
    & dropout rate & 0.3 & 0.3 & 0.6 \\\cline{2-5}
    & node embedding dimension & 64 & 64 &  64 \\\cline{2-5}
    & hidden layer dimension & 64 & 64 & 64 \\\cline{2-5}
    & readout embedding dimension & 64 & 64 & 64 \\\cline{2-5}
    & graph embedding dimension & 64 & 64 & 64 \\\cline{2-5}
    & attention heads & None & 2 &  None \\\cline{2-5}
    & alpha & None & 0.2 &  None \\\cline{2-5}
    \hline
    \multirow{10}{*}{QM9}
  & MAE Score & 21.075  & 23.173  & \textbf{19.167} \\\cline{2-5}
     & number of clients & 8  & 8 & 8 \\\cline{2-5}
    & learning rate & 0.0015 & 0.00015  & 0.15 \\\cline{2-5}
    & dropout rate & 0.2 & 0.5 & 0.3 \\\cline{2-5}
    & node embedding dimension & 64 & 256 &  64 \\\cline{2-5}
    & hidden layer dimension & 64 & 128 & 64 \\\cline{2-5}
    & readout embedding dimension & 64 & 256 & 64 \\\cline{2-5}
    & graph embedding dimension & 64 & 64 & 64 \\\cline{2-5}
    & attention heads & None & 2 &  None \\\cline{2-5}
    & alpha & None & 0.2 &  None \\\cline{2-5}
    \hline
    \end{tabular}
    \label{tab:ESOL_FreeSolv_Lipo_fed}
\end{table}

\begin{table}[h!]
\label{tab:subgraphparams}
    \small
    \centering
    \caption{Hyperparameters for Subgraph-Level Centralized Link Prediction Task}
    \begin{tabular}{ |c|c|c|c|c| } 
    \hline
    \textbf{Dataset} & \textbf{Score \& Parameters} & \textbf{GCN } & \textbf{GAT } & \textbf{GraphSAGE } \\
    \hline
     \multirow{8}{*}{Ciao}
    & mean absolute error & \textbf{0.8167} & 0.8214  & 0.8231 \\\cline{2-5}
    & mean squared error & \textbf{1.1184} & 1.1318  & 1.1541 \\\cline{2-5}
    & root mean squared error & \textbf{1.0575} & 1.0639  & 1.0742 \\\cline{2-5}
    & communication round & 100 & 100  & 50 \\\cline{2-5}
    & learning rate & 0.01 & 0.001 & 0.01 \\\cline{2-5}
    & node embedding dimension & 64 & 64 & 64 \\\cline{2-5}
    & hidden layer dimension & 32 & 32 & 16 \\\cline{2-5}
    \hline    
    \multirow{8}{*}{Epinions}
    & mean absolute error & \textbf{0.8847} & 0.8934  & 1.0436 \\\cline{2-5}
    & mean squared error & \textbf{1.3733} & 1.3873  & 1.8454 \\\cline{2-5}
    & root mean squared error & \textbf{1.1718} & 1.1767  & 1.3554 \\\cline{2-5}
    & communication round & 50 & 50  & 100 \\\cline{2-5}
    & learning rate & 0.01 & 0.01 & 0.01 \\\cline{2-5}
    & node embedding dimension & 64 & 64 & 64 \\\cline{2-5}
    & hidden layer dimension & 32 & 32 & 64 \\\cline{2-5}
    \hline    
    \end{tabular}
    \label{tab:subgraph_recsys}
\end{table}

\begin{table}[h!]
\label{tab:fedsubgraph-params}
    \small
    \centering
    \caption{Hyperparameters for Subgraph-Level Federated Link Prediction Task}
    \begin{tabular}{ |c|c|c|c|c| } 
    \hline
    \textbf{Dataset} & \textbf{Score \& Parameters} & \textbf{GCN + FedAvg} & \textbf{GAT + FedAvg} & \textbf{GraphSAGE + FedAvg} \\
    \hline
    \hline   
     \multirow{10}{*}{Ciao}
    & mean absolute error & 0.7995 & \textbf{0.7987}  & 0.8290 \\\cline{2-5}
    & mean squared error & \textbf{1.0667} & 1.0682  & 1.1320 \\\cline{2-5}
    & root mean squared error & \textbf{1.0293} & 1.0311  & 1.0626 \\\cline{2-5}
    & communication round & 100 & 100  & 50 \\\cline{2-5}
    & local epochs & 5 & 2  & 5 \\\cline{2-5}
    & number of clients & 8 & 8 & 8 \\\cline{2-5}
    & learning rate & 0.01 & 0.001 & 0.01 \\\cline{2-5}
    & node embedding dimension & 64 & 64 & 64 \\\cline{2-5}
    & hidden layer dimension & 32 & 32 & 32 \\\cline{2-5}
    \hline    
    \multirow{10}{*}{Epinions}
    & mean absolute error & 0.9033 & \textbf{0.9032}  & 0.9816 \\\cline{2-5}
    & mean squared error & 1.4378 & \textbf{1.4248}  & 1.6136 \\\cline{2-5}
    & root mean squared error & 1.1924 & \textbf{1.1882}  & 1.2625 \\\cline{2-5}
    & communication round & 100 & 50  & 100 \\\cline{2-5}
    & local epochs & 2 & 1  & 2 \\\cline{2-5}
    & number of clients & 8 & 8 & 8 \\\cline{2-5}
    & learning rate & 0.01 & 0.001 & 0.01 \\\cline{2-5}
    & node embedding dimension & 64 & 64 & 64 \\\cline{2-5}
    & hidden layer dimension & 64 & 64 & 64 \\\cline{2-5}
    \hline    
    \end{tabular}
    \label{tab:subgraph_recsys_fed}
\end{table}

\begin{table}[h!]
\label{tab:fedreg-node-params}
    \small
    \centering
    \caption{Hyperparameters for Node-Level Federated Node Classification Task}
    \begin{tabular}{ |c|c|c|c|c| } 
    \hline
    \textbf{Dataset} & \textbf{Parameters} & \textbf{GCN + FedAvg} & \textbf{GAT + FedAvg} & \textbf{GraphSAGE + FedAvg} \\
    \hline
    \hline    
    \multirow{5}{*}{CORA}
     & Micro F1 score & 0.8549 & diverge  & \textbf{0.9746} \\\cline{2-5}
     & number of clients & 10 & 10 & 10 \\\cline{2-5}
    & learning rate & 0.001 & 0.001  & 0.001 \\\cline{2-5}
    & dropout rate & 0.5 & 0.5  & 0.5 \\\cline{2-5}
    & local epochs & 5 & 5 & 5 \\\cline{2-5}
    & hidden layer dimension & 128 & 128 & 128 \\\cline{2-5}
    \hline    
    \multirow{5}{*}{Citeseer}
     & Micro F1 score & 0.9743 & 0.9610  & \textbf{0.9854} \\\cline{2-5}
    & learning rate & 0.001 & 0.01  & 0.001 \\\cline{2-5}
    & dropout rate & 0.5 & 0.5  & 0.5 \\\cline{2-5}
    & local epochs & 5 & 3 & 5 \\\cline{2-5}
    & hidden layer dimension & 128 & 128 & 128 \\\cline{2-5}
    \hline
    \multirow{5}{*}{PubMed}
     & Micro F1 score & 0.9191 & 0.8557  & \textbf{0.9761} \\\cline{2-5}
     & number of clients & 10 & 10 & 10 \\\cline{2-5}
    & learning rate & 0.001 & 0.001  & 0.001 \\\cline{2-5}
    & dropout rate & 0.5 & 0.5  & 0.5 \\\cline{2-5}
    & local epochs & 5 & 5 & 5 \\\cline{2-5}
    & hidden layer dimension & 128 & 128 & 128 \\\cline{2-5}
    \hline
    \multirow{5}{*}{DBLP}
     & Micro F1 score & 0.9088 & 0.8201  & \textbf{0.9749} \\\cline{2-5}
     & number of clients & 10 & 10 & 10 \\\cline{2-5}
    & learning rate & 0.001 & 0.001  & 0.001 \\\cline{2-5}
    & dropout rate & 0.5 & 0.5  & 0.5 \\\cline{2-5}
    & local epochs & 5 & 5 & 5 \\\cline{2-5}
    & hidden layer dimension & 128 & 128 & 128 \\\cline{2-5}
    \hline
    
    \end{tabular}
    \label{tab:node_level_fed_ego}

\end{table}

\clearpage
\subsection{Evaluation Metrics}
\label{app:experiments:metric}
Current metrics supported in \ours include:
\begin{myitemize}
\item \textbf{Graph Classification}: ROC-AUC (Area Under the Curve - Receiver Operating Characteristics) is a well-used classification metric to evaluate the performance at various thresholds.
\item \textbf{Graph Regression}: For this task, we chose RMSE (Root Mean Squared Error). However, when train and test distributions differ, it is more suitable to use metrics such as MAPE (Mean Absolute Percentage Error).
\item \textbf{Node Classification}: In ego-network node-level FL task, we use F1 score as the metric because F1 score is better than the accuracy metric when imbalanced class distribution exists.
\item\textbf{Link prediction (Recommendation Systems)}: Specifically for recommendation systems, we can treat it as a regression problem. Thus, it is possible to use well-known metrics such as MAE (Mean Absolute Error), MSE (Mean Squared Error), RMSE (Root Mean Squared Error). Widely used ranking based metrics can also be applied such as DCG (Discounted Cumulative Gain) and NDCG (Normalized Discounted Cumulative Gain),.
\item\textbf{Relation Prediction (Knowledge Graphs)}: Besides accuracy based metrics such as precision, recall and f1 score, ranking based metrics are also applied to relation type prediction on knowledge graphs such as MRR (Mean Reciprocal Rank) \cite{MRR} and HR (Hit Ratio).
\end{myitemize}


In addition to the metrics described, \ours users can add their custom metrics as well. As a future work, we plan to analyze these metrics' representative capacity on the FL performance.

\subsection{More Experimental Results}
\label{app:experiments:more}
Aside from additional results for graph-level FL, we will further provide more live results and reports in our project website \url{https://FedML.ai/FedGraphNN}. We hope these visualized training results can be a useful reference for future research exploration.

\noindent\begin{minipage}{0.5\textwidth}
\centering
\resizebox{\textwidth}{!}{
{\includegraphics[width=0.650\textwidth]{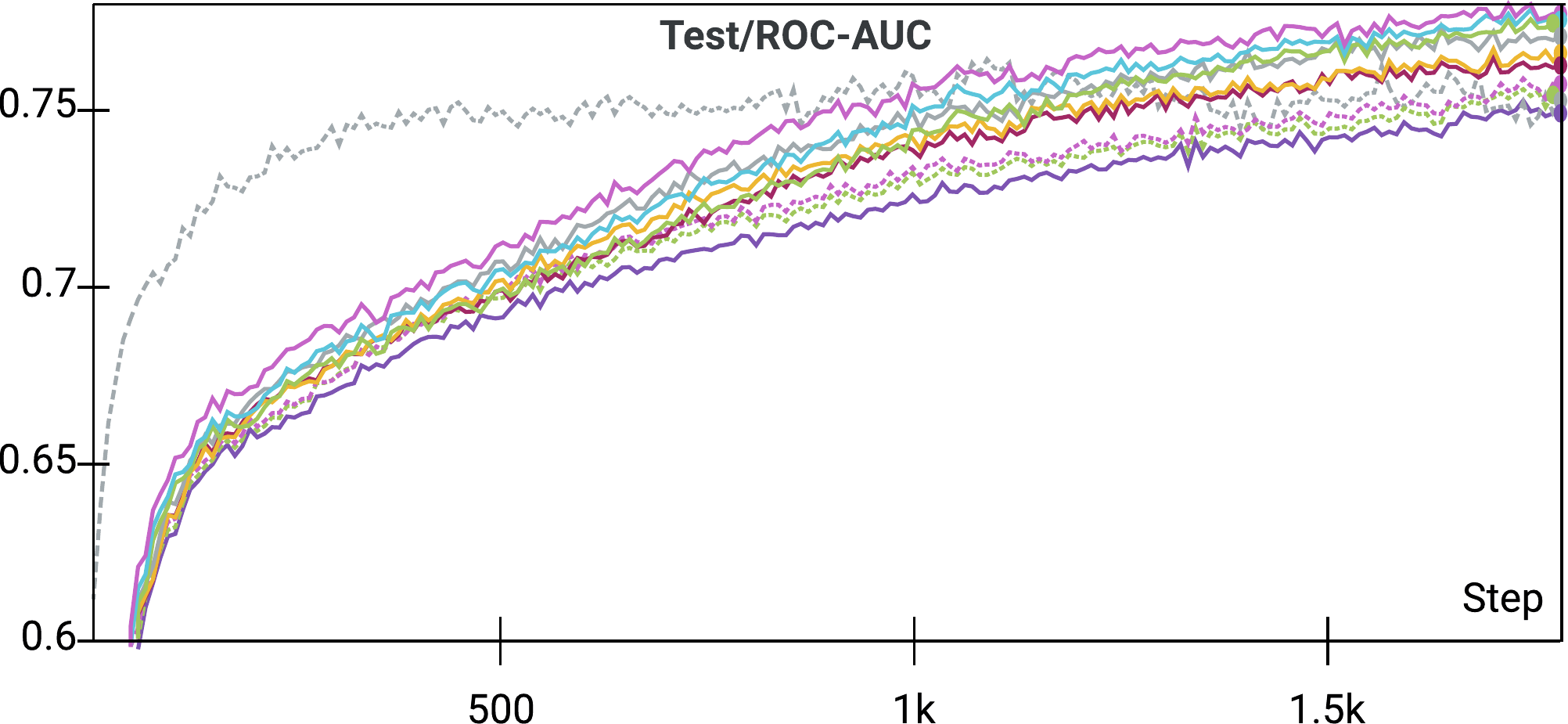}}}
\captionsetup{font=footnotesize,type=figure}
\captionof{figure}{Tox21: test score during sweeping}
\label{fig:tox21-sweep}
\end{minipage}\hfill
\begin{minipage}{0.5\textwidth}
\centering
\resizebox{\textwidth}{!}{
{\includegraphics[width=0.65\textwidth]{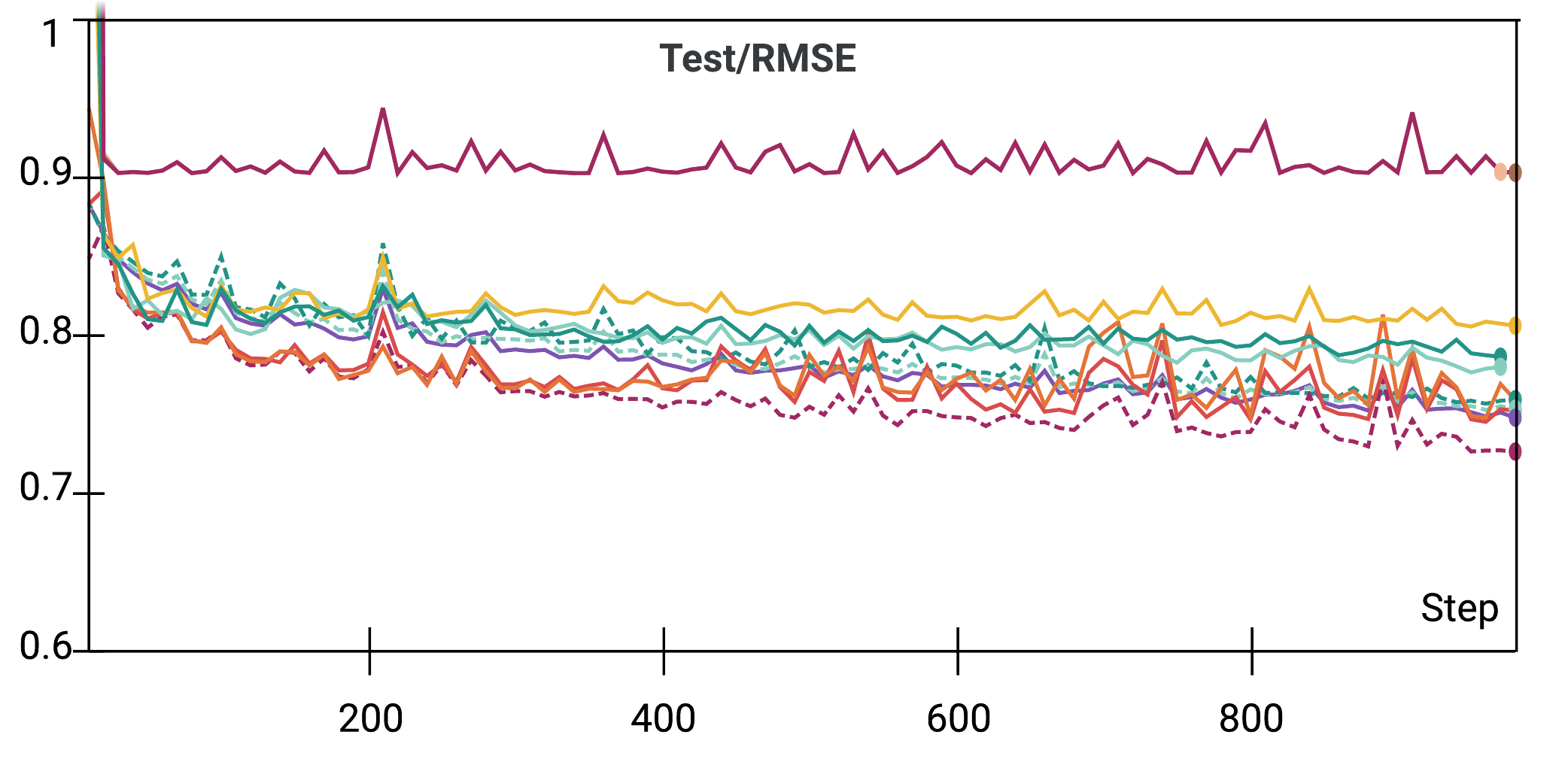}}}
\captionsetup{font=footnotesize,type=figure}
\captionof{figure}{hERG: test score during sweeping}
\label{fig:herg-sweep}
\end{minipage}

\begin{table*}[htbp]
\centering
\caption{Performance of graph regression in the graph-level FL setting (\#clients=4, MAE for QM9).}
\resizebox{\textwidth}{!}{
    \begin{tabular}{l  c c c c c   }
    \toprule
     Metric & \multicolumn{5}{c}{RMSE}  \\ 
     \cmidrule(lr){2-6}
     Method & \textsc{FreeSolv} & \textsc{ESOL} & \textsc{Lipo} & \texttt{hERG} & \textsc{QM9}  \\
      & \textsc{$\alpha = 0.2$} & \textsc{$\alpha = 0.5$} & \textsc{$\alpha = 0.5$} & \textsc{ $\alpha = 3$} & \textsc{$\alpha = 3$}   \\
     \midrule
        MoleculeNet Results & 1.40±0.16 & 0.97±0.01 & 0.655±0.036 & \texttt{DNE} & 2.35  \\
      \midrule
        GCN (centralized) & 1.5787 & 1.0190 & 0.8518 & 0.7257 & 14.78  \\
     GCN (FedAvg) & 2.7470 & 1.4350 & 1.1460 & 0.7944 & 21.075  \\ 
         \midrule
        GAT (Centralized) & 1.2175 & 0.9358 & 0.7465 & 0.6271 & 12.44  \\
     GAT (FedAvg) & 1.3130 & 0.9643 & 0.8537 & 0.7322 & 23.173 \\ 
       \midrule
        GraphSAGE (centralized) & 1.3630 & 0.8890 & 0.7078 &0.7132 & 13.06  \\
     GraphSAGE (FedAvg) & 1.6410 & 1.1860 & 0.7788 & 0.7265 & 19.167 \\ 
    \bottomrule
    \end{tabular}
}
\label{tab:graphlevel-reg}
\end{table*}

\end{document}